\documentclass[10pt,twocolumn]{IEEEtran}
\usepackage{graphicx}
\usepackage{amssymb}
\usepackage{epstopdf}
\usepackage{algorithmic}
\usepackage{subfigure}
\usepackage{amsthm}
\usepackage{marginnote}
\usepackage{marginnote}

\usepackage{color}
\definecolor{darkred}{rgb}{0.5,0,0}
\definecolor{darkgreen}{rgb}{0,0.5,0}
\definecolor{darkblue}{rgb}{0,0,1}
\usepackage[colorlinks]{hyperref}
\hypersetup{ colorlinks,
linkcolor=darkblue,
filecolor=darkgreen,
urlcolor=darkred,
citecolor=darkblue }
\usepackage{cite}

\newcommand{\mbf}[1]{{\mathbf #1}}

\newcommand{\bkt}[1]{\left( #1\right)}

\newcommand{\abs}[1]{\left| #1\right|}

\title{Deformation corrected compressed sensing (DC-CS): a novel framework for accelerated \textcolor{black}{dynamic} MRI}
\author{Sajan Goud Lingala,~\IEEEmembership{Student Member,~IEEE}, Edward DiBella, \IEEEmembership{Member,~IEEE} and Mathews~Jacob,~\IEEEmembership{Senior Member,~IEEE} 
\thanks{Copyright (c) 2010 IEEE. Personal use of this material is permitted. However, permission to use this material for any other purposes must be obtained from the IEEE by sending a request to pubs-permissions$@$ieee.org. \newline Sajan Goud Lingala is with the Department of Biomedical Engineering, The University of Iowa, IA, USA (email: sajangoud-lingala@uiowa.edu). Edward DiBella is with the Department of Radiology, University of Utah, Utah, USA. \newline Mathews Jacob is with the Department of Electrical and Computer Engineering, The University of Iowa, IA, USA.\newline This work is supported by grants  NSF CCF-0844812, NSF CCF-1116067, NIH 1R21HL109710-01A1, and AHA 12 PRE11920052.}}

\begin{document}
\maketitle

\begin{abstract}
We propose a novel deformation corrected compressed sensing (DC-CS) framework to recover \textcolor{black}{contrast enhanced}  dynamic magnetic resonance images from undersampled measurements. We introduce a formulation that is capable of handling a wide class of sparsity/compactness priors on the deformation corrected dynamic signal. 
In this work, we consider example compactness priors such as sparsity in temporal Fourier domain, sparsity in temporal finite difference domain, and nuclear norm penalty to exploit low rank structure.  Using variable splitting, we decouple the complex optimization problem to simpler and well understood sub problems; the resulting algorithm alternates between simple steps of shrinkage based denoising, deformable registration, and a quadratic optimization step. Additionally, we employ efficient continuation strategies to reduce the risk of convergence to local minima. \textcolor{black}{The decoupling enabled by the proposed scheme enables us to apply this scheme to contrast enhanced MRI applications.}  Through experiments on numerical phantom and in vivo myocardial perfusion MRI datasets, \textcolor{black}{we observe superior image quality of the proposed DC-CS scheme in comparison to the classical k-t FOCUSS with motion estimation/correction scheme, }and demonstrate reduced motion artifacts over classical compressed sensing schemes that  utilize the compact priors on the original deformation un-corrected signal.
\end{abstract}

\section{Introduction}

\begin{figure*}[t!]
\centering
{\includegraphics[width=0.91\textwidth]{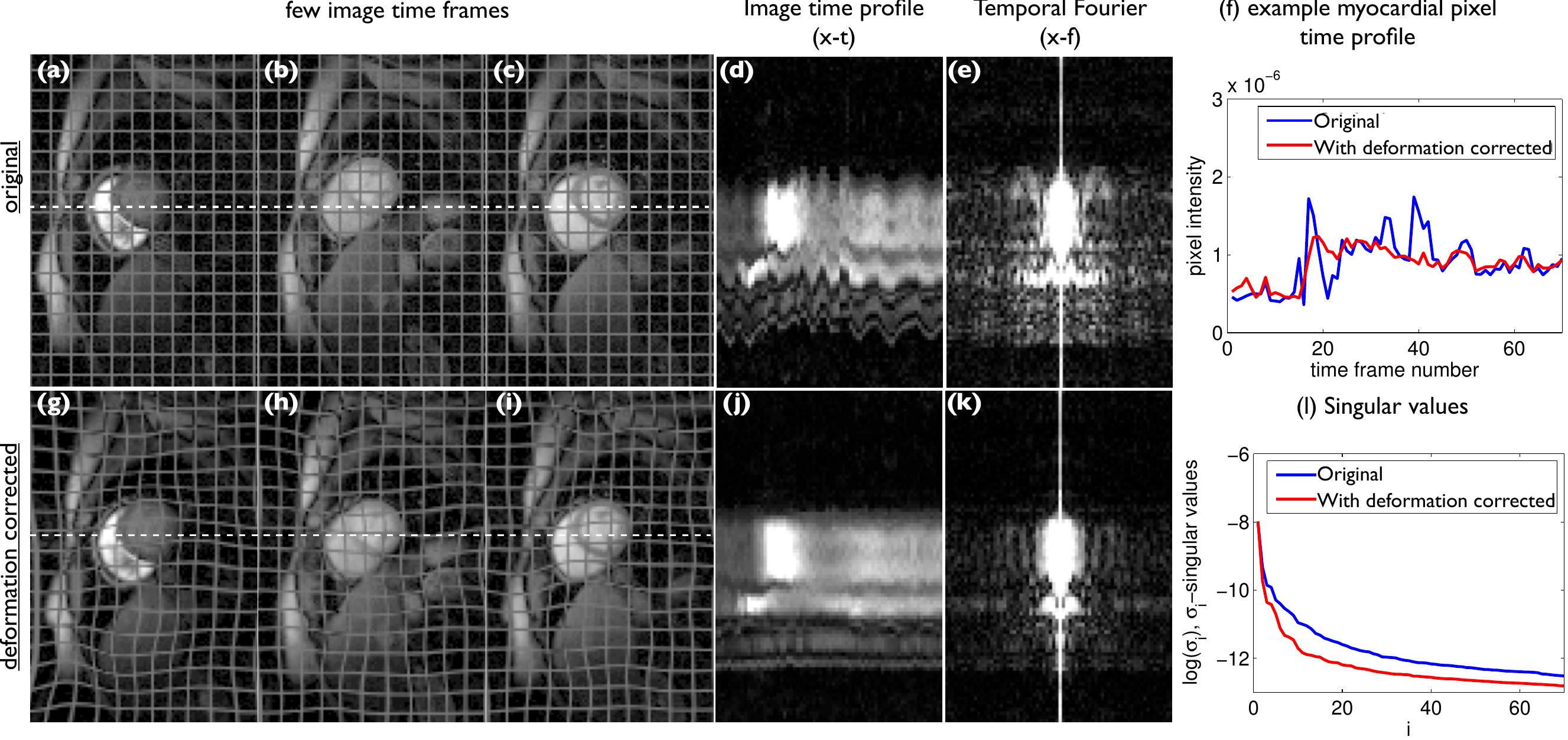} }

\caption{Free breathing myocardial perfusion MRI data representation in transform domains with and without deformation correction: We show a few example dynamic frames from a myocardial perfusion MRI dataset that contains considerable interframe motion in (a-c). The corresponding deformation/motion  corrected dataset is shown in (g-i). A mesh pattern is superimposed on the images for better visualization of the deformation. The image time profiles of the original and deformation due to motion corrected datasets along the white dotted lines on the frames are shown in (d,j); the ripples in (d) correspond to the motion largely due to breathing. The corresponding profile in (j) show that  profile is largely free of the ripples. From (e) and (k), it can be seen that the deformation corrected dataset has a sparser representation in the temporal Fourier domain, compared to the original data as the temporal harmonics corresponding to the motion are compensated in (k). From (f), it can be seen that the pixel time profiles are more piece-wise smooth with the deformation corrected data compared to the original, hence the former has more sparse temporal gradients. It is seen from (l) that the number of significant singular values are reduced in the deformation corrected dataset compared to the original. }
\label{motionschematic}
\end{figure*}

\textcolor{black}{Dynamic magnetic resonance imaging (DMRI) involves imaging objects that are evolving in time, and is central to several clinical exams including cardiovascular, pulmonary, abdominal, brain, and vocal tract imaging}. \textcolor{black}{DMRI} often suffers from compromises in image quality due to the slow acquisition nature of MRI. For instance, good spatio-temporal resolution, extended slice coverage, and high signal to noise ratio are required for accurate quantification of myocardial perfusion MRI data. However, imaging at Nyquist k-space sampling rate often results in severe compromises in spatio-temporal resolution and slice coverage \cite{kellman2007imaging}. Classical approaches to overcome these challenges include parallel imaging \cite{griswold2002generalized, plein2005coronary}, and their combination with $k-t$ spatio-temporal models \cite{tsao2003k, plein2007dynamic, sharif2010patient, di2003comparison, gupta2001dynamic, liang2007spatiotemporal, pedersen2009k, tsao2012mri}. Recently, several authors have proposed compressed sensing (CS) schemes that capitalize on the compactness/sparsity of the signal representation in appropriate transform domains. For example, sparsity of the temporal Fourier transform \cite{otazo2010combination} and temporal finite differences  \cite{adluru2007temporally} have been exploited in the context of myocardial perfusion MRI.  More recently, matrix recovery schemes utilizing the  linear dependancies of pixel time profiles using low rank image priors have been proposed \cite{lingala2011accelerated, haldar2010spatiotemporal, trzasko2011local}. While all of these methods demonstrate successful recovery when the inter frame motion is minimal, the main challenge is the sensitivity of these methods to large inter frame motion. Specifically, the compactness of the signal representation decreases with inter frame motion, thus restricting the maximum possible acceleration  (see Fig.\ref{motionschematic} for a demonstration); the reconstructions often suffer from temporal blurring and motion related artifacts at high acceleration factors.

 
In this work, we introduce a general framework to minimize the sensitivity of compressed sensing and low rank matrix recovery schemes to inter frame motion. We jointly estimate the dynamic images and  inter frame motion, which is modeled as an elastic deformation, from the undersampled data. Rather than assuming compactness of the original signal, we assume the deformation corrected signal to have a compact representation. The proposed approach enables us to use arbitrary signal priors (e.g. sparsity in specified transform domain, low-rank property, patch based low-rank priors) in the reconstruction; the appropriate method could be chosen depending on the specified application. We introduce an efficient variable splitting framework with continuation to decouple the problem into three simpler and well-understood sub-problems. We alternate between (a) a shrinkage based denoising step (b) a deformable registration step, and (c) a quadratic optimization step. The deformable registration scheme aims to \textcolor{black}{register each frame in the dataset to a corresponding frame with similar contrast in the motion-compensated dataset}. Hence, simpler least squares difference metrics are sufficient for the registration algorithm, even when the image contrast changes with time, such as in  dynamic contrast enhanced MRI. The presence of the global energy function enables us to design appropriate continuation strategies to reduce the \textcolor{black}{risk of} convergence of the algorithm to local minima. {A preliminary version of this work was published as a conference proceeding in \cite{lingala2011unified}. }

In this paper, we demonstrate the utility of the \textcolor{black}{proposed} formulation \textcolor{black}{in the context of myocardial perfusion MRI}. We consider example compactness priors such as sparsity in \textcolor{black}{the} temporal Fourier domain, sparsity in temporal finite difference domain, and nuclear norm penalty to exploit low rank structure. While the proposed formulation is capable of handling  non-convex variants of these priors ($\ell_{p}$, Schatten $p$-norms $(p<1)$) and their combinations (e.g. \cite{lingala2011accelerated}), these variants are beyond the scope of this paper. Our experiments  demonstrate the utility of the proposed scheme in improving the reconstructions in terms of reduced motion artifacts and better spatio-temporal fidelity compared to schemes that utilize the compact priors on the original deformation un-corrected signal. 

The proposed unified energy minimization formulation is conceptually related  to the elegant work by Fessler et. al. in the context of reconstructing a static image of a moving organ from its measurements \cite{fessler2010optimization}. \textcolor{black}{To our knowledge, the earliest work in static MRI reconstruction of a moving organ was by \cite{odille2008generalized}, and the earliest work of joint image reconstruction and motion compensation in dynamic imaging was in positron emission tomography  \cite{jacobson2003joint,reyes2007model}. The dynamic imaging approaches rely on the registration of each frame of the dataset to a reference frame.} Similar strategies were recently introduced for CINE MRI. For example, 
 Jung et. al, have extended the $k-t$ FOCUSS scheme with motion estimation and compensation for cardiac cine MRI \cite{jung2010motion}. This scheme approximate the dynamic images as the deformation of fully sampled reference frames. The residuals are then reconstructed from under-sampled $k$-space data using $k-t$ FOCUSS. Similarly, Asif et al. in \cite{asif2012motion} demonstrated a scheme for cardiac cine MRI that alternate between motion estimation and motion constrained CS reconstruction problems; the model estimates the inter frame motion by registering neighboring frames in the dynamic sequence. Other motion compensated schemes customized to free breathing cardiac cine and delayed enhancement MRI have also been introduced \cite{kellman2009high, kellman2008fully, usman2012motion, hansen2012retrospective, ledesma2007motion}. Unlike cine MRI, the contrast of the dynamic perfusion images are significantly different from the reference images. Hence, the subtraction of the deformed reference image may not generate sparse residuals. Moreover,  more complex similarity measures and models may be needed for the registration  as image contrast varies significantly across time frames\cite{wollny2008new, wollny2012automatic, stegmann2005unsupervised, adluru2006model, wollny2011nonrigid, khalifa2012new}. In the context of contrast enhanced DMRI,  Pederson et. al proposed to unify the reconstruction of the images and the motion compensation into a single algorithm \cite{pedersen2010unifying}.  They represented the contrast variations using a parametric perfusion model, while the deformation due to motion was modeled as a modulation of a 2-D displacement field, which is estimated from two images acquired at end inspiration and end expiration. The fewer degrees of freedom in this model may be restrictive in practical perfusion imaging applications. In contrast, the proposed model is considerably less constrained than the parametric scheme used in \cite{pedersen2010unifying}. The proposed approach does not require fully sampled pre-scans or navigators for motion estimation. In addition, since we do not model the dynamic frames as deformations of pre-contrast reference images, our approach is robust to contrast variations due to bolus passage. Otazo et al. in \cite{otazo2011combination} partially corrected for the motion in myocardial perfusion MRI using a rigid deformation model, where all the frames from a preliminary CS reconstruction were mapped to a single fully sampled reference image to estimate the motion.  \textcolor{black}{The proposed scheme can be viewed as a systematic alternative for patch based low-rank methods that were recently introduced \cite{chen2013motion,yoon2014motion}. These methods cluster similar patches from adjacent frames and enforce low-rank penalties on them. The main challenge with these schemes is the extensive book-keeping to ensure that all patches are in some cluster. }We have also investigated the use of a model-based registration method to handle simultaneous motion and changes in contrast \cite{adluru2013reconstruction}.

The rest of the paper is organized as described. In sections II, III, we describe the formulation of the DC-CS problem, the proposed variable splitting framework, and the resulting optimization algorithm. In sections IV, V, we present our experimental evaluation and discussion on the feasibility of the DC-CS algorithm to improve reconstruction quality of free breathing myocardial perfusion MRI data based on retrospective resampling experiments on fully sampled numerical phantom and in-vivo datasets. We also demonstrate the feasibility in improving free breathing reconstructions of radial data from a subject imaged during adenosine stress perfusion. 

\section{Deformation corrected compressed sensing (DC-CS)}

\subsection{Dynamic image acquisition}
The main objective of this paper is to recover the dynamic dataset $f(\mathbf x, t)$ from its undersampled Fourier noisy measurements $b(\mathbf k_{i},t_{i})$. Here, $\mathbf x$ is the spatial variable and $t$ denotes time. The measurement process in dynamic MRI can be modeled as: 
\begin{eqnarray}
\label{Fourier}
\mathbf b_{i} =\int_{\mbf x} f(\mbf x,t_{i}) \exp\bkt{-j \mbf k_{i}^{T} \mbf x}d\mbf x + \mathbf n_{i}; \;\;i=0,.,s-1;
\end{eqnarray}
Here, $(\mbf k_{i},t_{i})$ indicates the $i^{\rm th}$ sampling location in $k-t$ space. We denote the set of sampling locations as $\Xi=\{(\mbf k_{i},t_{i}),i=0,..,s-1\}$. The above expression can be rewritten in the vector form as 
\begin{eqnarray}
\label{measurementmodel}
\mbf b={\cal A}(f)+\mbf n,
\end{eqnarray}
 where ${\cal A}$ is an operator that evaluates the Fourier Transform on the sampling locations specified in $\Xi$.

\subsection{DC-CS model}
The pixel time profiles in \textcolor{black}{myocardial perfusion MRI} are highly structured in the presence of perfect gating and breath-holding. As demonstrated in Fig. \ref{motionschematic}, penalties such as temporal Fourier sparsity (to exploit low temporal bandwidth), temporal total variation penalty (to exploit  smooth pixel time profiles) or low-rank penalties (to exploit the redundancy between the pixel time profiles) can be used to make the recovery from under sampled data well posed. However, the compactness of the signal representations will be considerably disturbed in the presence of  inter frame motion, which can arise due to breathing or inconsistent gating (see Fig. \ref{motionschematic}); due to which, the performance of the above schemes will be heavily compromised. 

We propose to overcome the above limitation by assuming the deformation corrected dataset ${\cal T}_{\theta} \cdot f$ to be compact/sparse.  Note from Fig. \ref{motionschematic}.(f),(j),(k),(l) that the temporal variations of ${\cal T}_{\theta} \cdot f$ are considerably more structured than that of $f$\footnote{ \textcolor{black}{In figure 1, registration was performed on the free breathing dataset itself. Starting from the second frame, the deformations were obtained by matching the nth frame in the moving sequence to the (n + 1) th frame of the deformed scene by using the demons registration algorithm \cite{matlablinkdemons}}}. We simultaneously recover the deformation parameters $\theta(\mathbf x,t)$ and the dynamic images $f(\mathbf x,t)$ from under sampled data $\mathbf b(k, t)$ using the following minimization scheme:
\begin{eqnarray}\label{maincost}\nonumber
\{f^{*},\theta^{*}\} &=& \min_{f, \theta} \|{\cal A}(f)-\mathbf b\|_{2}^{2} + \lambda~\Phi\left({\cal T}_{\theta} \cdot f\right);\\
&&\qquad\mbox{ such that  } \theta \in \mbf \Theta
\end{eqnarray}
Here $\cal A$ is the Fourier sampling operator as defined in (\ref{measurementmodel}) and ${\cal T}_{\theta}$ is the non-rigid image warping operator; $\theta(\mathbf x,t)$ are the deformation parameters that describe pixel wise displacements due to motion, which are estimated from the under sampled data. The proposed scheme also yield ${\cal T}_{\theta(\mathbf x,t)}\cdot f$, which is the deformation corrected version of $f$, as a by-product. Note that (\ref{maincost}) reduces to the classical CS setting when ${\cal T}_{\theta} = {\cal I}$: the identity operator. We use bilinear interpolation to discretize the term ${\cal T}_{\theta}\cdot f$.

The regularization term in (\ref{maincost}) promotes the compactness/sparsity of the deformation corrected dataset ${\cal T}_{\theta}\cdot f$, rather than $f$. Here, $\Phi(\textcolor{black}{u})$ denotes an arbitrary prior to exploit the redundancy in the data; $\lambda$ is the corresponding regularization parameter. The main advantage of the proposed scheme is that it can be used with any spatio-temporal priors on the deformation corrected dataset. The specific priors can be chosen depending \textcolor{black}{on} the specific application. For example, we can choose $\Phi(\textcolor{black}{u})=\|{\cal F}_{t}\|_{\ell_{1}}$ or $\Phi(\textcolor{black}{u})=\|{\nabla}_{t}\|_{\ell_{1}}$, where ${\cal F}_{t}$ and $\nabla_{t}$ are the temporal Fourier transform and temporal gradient operator, respectively. Another alternative is to choose $\Phi(\textcolor{black}{u}) = \|\textcolor{black}{\mathbf U}\|_{*}$, the nuclear norm of the Casorati matrix \textcolor{black}{$\mathbf U$}  associated with \textcolor{black}{$u(\mathbf x,t)$} \cite{liang2007spatiotemporal, lingala2011accelerated}. This approach exploits the low-rank property of the deformation/motion corrected dataset, resulting from the similarity between the temporal profiles of the pixels \cite{lingala2011accelerated, haldar2010spatiotemporal, trzasko2011local}. The ability of the scheme to handle arbitrary image priors makes this approach drastically different from classical motion compensation schemes that register each frame of $f$ to a specific fully sampled frame.

\textcolor{black}{The deformation field in (\ref{maincost}) is assumed to be parametrically represented in terms of the parameters $\theta$. For example, $\Theta$ is the set of B-spline coefficients if a B-spline model is used to represent the deformation field as in \cite{kybic2003fast, chun2009simple}. In this case, the  spatial smoothness of the deformation map is controlled by the grid spacing of the B-spline map. The  spatial smoothness constraints can also be explicitly imposed using regularization constraints on the deformation field as in \cite{chefd2002flows}. Our approach is closely related to  \cite{chefd2002flows} as explained in detail in Section \ref{mest}.}


\section{DC-CS: Optimization algorithm}
We propose to use a variable splitting approach \cite{afonso2010fast,ramani2012splitting} to  decouple the original problem in (\ref{maincost}) to simpler sub-problems. We start by splitting the deformation term from the prior by introducing an auxiliary variable $g$. This enables us to reformulate the problem in (\ref{maincost}) to the following constrained optimization scheme: 
\[
\{f,\theta\}^{*} = \arg \min_{f, \theta,g} \|{\cal A}(f)-\mathbf b\|_{2}^{2} + \lambda~ {\Phi\left(g\right)};\newline
\]
\begin{equation}\label{const}
\mbox{such that   }~{\cal T}_{\theta} \cdot f = g,~ \theta \in \mbf \Theta
\end{equation}
We solve (\ref{const}) using the penalty based method, where the first constraint in $(\ref{const})$ is relaxed and the corresponding quadratic violation is penalized as:
\[
\{f,\theta\}^{*} = \arg \min_{f, \theta,g} \|{\cal A}(f)-\mathbf b\|_{2}^{2} + \lambda \left({\Phi\left(g\right)} + \frac{\beta}{2}\| {\cal T}_{\theta} \cdot f - g\|_{2}^{2}\right);
\]
\begin{equation}
\label{penaltycost}
\mbox{such that}~ \theta \in \mbf \Theta
\end{equation}
Here, $\beta$ is the penalty parameter that enforces the constraint $({\cal T}_{\theta}\cdot f=g)$. When $\beta$ approaches $\infty$, the solution of $(\ref{penaltycost})$ tends to that of $(\ref{const})$, and equivalently $(\ref{maincost})$. However, the convergence of the algorithm will be slow when $\beta$ is very high and the algorithm may also be vulnerable to local minima. Hence, we will use a continuation strategy where $\beta$ is initialized with very low values and is gradually increased, as discussed in detail below. 

We use an \textcolor{black}{iterative} minimization scheme to minimize $(\ref{penaltycost})$ with respect to three variables $f, \theta, g$. Specifically, \textcolor{black}{the iterative algorithm}  alternates between the minimization of each variable, assuming the rest to be known. This approach results in the following sub problems.

\subsection{ $g$ sub-problem (\textcolor{black}{Spatio-temporal denoising/dealiasing to minimize residual motion})} 

With $f$ and $\theta$ fixed, the minimization of $(\ref{penaltycost})$ with respect to $g$ \textcolor{black}{at each step} is a denoising problem; it involves the \textcolor{black}{proximal mapping \cite{rockafellar1997convex}} 
of the deformation corrected dataset $\underbrace{{\cal T}_{\theta}\cdot f}_{q}$:

\begin{equation}
\label{gsub}
g^{*} = \min_{g}   \frac{2}{\beta}\,\Phi(g)+\| \underbrace{{\cal T}_{\theta} \cdot f}_{q} - g\|_{2}^{2};
 \end{equation}

 \textcolor{black}{Note that the above implies $q$ is close enough to $g$, while having a small cost $\phi(g)$. In many cases, we can find the $g^{*}$ analytically, as seen in the below section. }

The amount of regularization is specified by the parameter $\beta$. Specifically, when $\beta$ is small, the optimal $g^{*}$ is a highly smoothed version of $q$.
\textcolor{black}{At each step, the proximal mapping smooths out the residual motion induced rapid temporal variations in the deformation corrected dataset $({\cal T}_{\theta}\cdot f)$, yielding the next iterate of the motion compensated dataset $(g^{*})$. The alternation of the subproblems (\ref{gsub}),(\ref{fsub}) and (\ref{thetasub}) results in the joint estimation of the dynamic images $(f)$ and the deformation map $(\theta)$. 
}

 
 \vspace{1em}
\subsubsection{{Sparsity penalty in the temporal Fourier transform domain}}
We denote \textcolor{black}{$\hat g = {\cal F}_{t}g$} as the temporal Fourier Transform of $g$. Using Parseval's theorem, we rewrite (\ref{gsub}) as
\begin{equation}
\hat g = \min_{\hat g }   \frac{2}{\beta}\, \|\hat g \|_{\ell_{1}}+\| \hat {q} - \hat g\|_{2}^{2}; 
\end{equation} 
Note that this is a standard $\ell_{1}$ proximal mapping \cite{rockafellar1997convex}. The optimal $\hat g$ is obtained by the shrinkage of $\hat q$ \cite{donoho1995noising}:
\begin{equation}
g = \frac{{\cal F}_{t}(q)}{\abs{{\cal F}_{t}(q)}}\left(\abs{{\cal F}_{t}(q)}-\frac{2}{\gamma\beta}\right)_{+},
 \label{xfsub}
 \end{equation}
 where $+$ is the shrinkage operator defined as:
 \begin{equation}
\label{l1shrinkage}
x_{+} = \left\{
\begin{array}{ccc}
x &\mbox{  if  }   & x \geq 0   \\
0  & \mbox{  else } 
\end{array}
\right..
 \end{equation}

\subsubsection{{Nuclear norm penalty}}
When $\Phi(g) = \|\mathbf G\|_{*}$, we rewrite (\ref{gsub}) as
\begin{equation}
\label{shrink2}
\mathbf G^{*} = \min_{\mathbf G }   \frac{2}{\beta}\, \|\mathbf G  \|_{*}+\| \mathbf Q - \mathbf G\|_{F}^{2},
\end{equation} 
where $\mathbf G, \mathbf Q$ are the Casorati matrices respectively associated with $g(\mathbf x,t)$ and $q(\mathbf x,t)$. The solution for the above proximal mapping is obtained analytically as \cite{cai2010singular}
\begin{equation}
\label{nuclearnormshrinkage}
\mathbf G^{*} = \mathbf U \left(\mathbf S - \frac{2}{\gamma\beta}\, \mathbf I\right)_{+}\mathbf V^{*},
\end{equation}
where $\mathbf U\mathbf S\mathbf V^{*}=\mathbf Q$ is the singular value decomposition of $\mathbf Q$, and $+$ is the shrinkage operator  as defined in $(\ref{l1shrinkage})$.
 \subsubsection{{Temporal total variation (TV) penalty}}
 When the penalty is chosen as the temporal total variation criterion, (\ref{gsub}) reduces to  
 \begin{equation}
 \label{tvdenoise}
g^{*} = \min_{g}  ~\| {q} - g\|_{2}^{2} + \frac{2}{\beta}\, \|\nabla_{t} g\|_{\ell_{1}} 
\end{equation}
 This TV denoising problem does not have an analytical solution unlike the above cases. Hence, we solve them using the fast TV denoising algorithm (FTVd) \cite{yang2010fast}. Briefly, this approach uses a splitting of (\ref{tvdenoise}), followed by an alternating algorithm to solve for $g^{*}$.

\subsection{{$f$ sub-problem (reconstruction update)}}

Assuming $\theta$ and $g$ fixed in $(\ref{penaltycost})$, the minimization with respect to $f$ reduces to:
\begin{equation}
\label{fsub}
\min_{f}\|{\cal A}(f)-\mathbf b\|_{2}^{2}+\frac{\lambda\beta}{2}\|{\cal T}_{\theta}\cdot f-g\|_{2}^{2};
\end{equation}
We solve this quadratic problem using the conjugate gradient (CG) algorithm. \textcolor{black}{Note that when $\beta$ approaches infinity, $g$ tends to ${\cal T}_{\theta}\cdot f$, and  (\ref{const}) tends to  (\ref{maincost}), and the solution to the $f$ subproblem in (\ref{fsub}) tends to the solution of (\ref{maincost}), which is the original problem we seek to solve.}

\subsection{ {${\cal \theta}$ sub-problem (Motion estimation)}} 
\label{mest}
Assuming the variables $f$ and $g$ in $(\ref{penaltycost})$ to be known, we solve for the motion parameters as: 
\begin{equation}
\label{thetasub}
 \min_{\theta}  \|{{\cal T}_{\theta} \cdot f} - g\|_{2}^{2}, ,~ \mbox{ such that } \theta \in \mbf \Theta
 \end{equation}
This is a registration problem, where the dynamic scene $f(\mathbf x,t)$ is registered \textcolor{black}{frame by frame} with a reference scene $g(\mathbf x,t)$. Since the reference series $g$ is derived from the measurements itself (obtained from $(\ref{gsub})$), we do not need to acquire additional high resolution reference frames. In addition, the least squares similarity metric in (\ref{thetasub}) sufficient, even when the contrast of the images are changing as a function of time, because each frame of $g(\mathbf x,t)$ has the same contrast as  $f(\mathbf x,t)$. The temporal profiles of the reference dataset $g$ is significantly more smooth compared to $f$. This approach enables us to decouple the effects of smooth perfusion induced contrast changes and the more rapid changes resulting from respiratory motion. 

\textcolor{black}{The ideal deformable registration algorithm will directly minimize (\ref{thetasub}) with a smoothness constraint on the deformation field. In addition, the continuous deformable parameters should be represented using a differentiable signal model such as cubic B-spline representation \cite{kybic2003fast, chun2009simple}. In this paper, we use the demons algorithm \cite{thirion1998image,wang2005validation} to approximate (\ref{thetasub}). This is largely motivated by available  open source implementation of the demons algorithm \cite{matlablinkdemons}. We observe that updating the deformation map using the demons force field is a good approximation to the steepest descent minimization of (\ref{thetasub}), subject to elasticity regularization (see \cite{pennec1999understanding} for details). In addition, the comparison of the demons algorithm and the direct steepest descent minimization of (\ref{thetasub}) subject to elasticity regularization is reported to provide qualitatively similar results \cite{bro1997medical}. We consider the  demons two-dimensional  registration algorithm that corrects for in-plane breathing motion, although in theory a model to correct for out of plane motion could be considered, when dealing with three-dimensional datasets. The demons algorithm is an iterative scheme, where the displacement field \textcolor{black}{$\theta(\mathbf x,t_{i})$} is updated as $\theta_{n+1} = \theta_{n} + u_{n}$, where $u_{n}$ is the force field evaluated at the $n^{\rm th}$ iteration. We refer the reader to \cite{thirion1998image,wang2005validation} for the expression of the force field; the force field is derived from intensity differences in the reference and target images . As discussed previously, the smoothness of the deformation field is implicitly enforced by \textcolor{black}{spatial smoothing} the force field $u_{n+1}$ using an isotropic Gaussian filter ${\cal G}_{\sigma}(\mathbf x)$, where $\sigma$ is the standard deviation of the Gaussian kernel.  Note that $\sigma$ is the spatial smoothness parameter, and allows for spatially regularizing the deformation maps. }

\subsection{Continuation strategies to alternate between the sub-problems}

The simultaneous estimation of $f$ and $\theta$ according to $(\ref{maincost})$ is a non-convex optimization problem. The variable splitting strategy described above enabled the decomposition of the original problem into three simpler subproblems in $(\ref{gsub}),(\ref{thetasub}),(\ref{fsub})$. However, the algorithm is not guaranteed to converge to the global minimum of the optimization problem.  We now introduce continuation strategies to reduce the risk of convergence to local minima and to ensure fast convergence. This approach is analogous to coarse to fine refinement strategies that are typically used in many image processing applications.

\subsubsection{ {Continuation over the penalty parameter $(\beta)$}} The optimization problem specified by $(\ref{penaltycost})$ has to be solved with a large value of $\beta$ to ensure that the constraint in $(\ref{const})$ is satisfied. However, it is known that the resulting algorithm will have poor convergence properties when $\beta$ is set to be high, even when ${\cal T}= \cal I$  \cite{hu2012fast}.  Specifically, the shrinkage step in (\ref{gsub}) essentially yields $g \approx {\cal T}_{\theta}\cdot f$ when $\beta \rightarrow \infty$. By contrast, if a low value of $\beta$ is used, $g$ computed using (\ref{gsub}) is a heavily denoised version of ${\cal T}_{\theta}\cdot f$. 

We adapt the continuation scheme from \cite{hu2012fast,yang2010fast,yang2011alternating} to obtain faster convergence. \textcolor{black}{The use of continuation strategy can be thought of as a coarse to fine refinement scheme. Specifically, we start with a low value of $\beta$, when the the cost function is considerably more smooth than the original one. Once the algorithm has converged to the global minimum of this cost, we  increase $\beta$; the algorithm then converges to a local minimum of the new cost function, which is close enough to the global minimum of the simpler cost function (with smaller value of $\beta$). Thus, this continuation approach will encourage the convergence of the algorithm to the global minimum, analogous to multi-resolution strategies used in non-convex problems such as image registration. We emphasize that the reference scene  is a result of continuation over $\beta$. We do not adapt approaches such as augmented Lagrangian, and split-Bregman that use a fixed value of $\beta$, which may lose the property of gradual convergence, while solving non-convex problems.}

\subsubsection{ {Continuation over a deformation force strength parameter}}

The motion estimation problem in $(\ref{thetasub})$ is itself a non-convex problem. Additionally when iterated along with the reconstruction $(\ref{fsub})$ and denoising $(\ref{gsub})$ problems, there is a possibility that the estimated motion parameters could get stuck  in undesirable local minima. Registration schemes often rely on coarse to fine continuation strategies to overcome local minima problems and to improve convergence  \cite{wang2005validation, ledesma2007motion, kybic1999multiresolution}.

The demons implementation \cite{wang2005validation} also recommends a continuation on a force strength parameter $\alpha$ to speed up the convergence of the algorithm and to minimize local minima effects. The parameter $\alpha$ manipulates the force field such that it is sensitive to large deformations when $\alpha$ is small, while larger values of $\alpha$ makes the force field sensitive to finer deformations \cite{wang2005validation}. Based on this, we adapt a continuation strategy of correcting for bulk motion during the initial iterations by using a small value of $\alpha$. As the iterations proceed, we gradually increase $\alpha$, and correct for finer motion changes. Specifically, with an initial guess of $\theta=0$,  the registration sub-problem in $(\ref{thetasub})$ is solved in an outer loop starting with a small value of $\alpha$, and the solution in $\theta$ is refined by gradually incrementing $\alpha$ towards high values; After the first outer loop, the sub-problem in $(\ref{thetasub})$ is initialized with the motion estimates obtained from the previous iteration. We have observed that this continuation accelerates the convergence of the DC-CS scheme considerably. \textcolor{black}{Each time the demons algorithm is called, we run it for a maximum of 100 iterations before termination. }

The following pseudo code summarizes the continuation strategies that we adapt to solve the cost in $(\ref{penaltycost})$:

\begin{center}
\line(1,0){250}
\end{center}
 \textbf{Initialization}: $f=f_{init}, \theta = 0, \alpha>0, \beta>0$;\\
 \textbf{for } \mbox{out = 1 to} $\rm{Max_{outer \hspace{0.21em}iterations}}$  \\
\hspace*{3em}  \textbf{for } \mbox{in = 1 to} $\rm{Max_{inner \hspace{0.21em}iterations}}$ \\ 
       \hspace*{6em} $g_{n}\leftarrow$$(\ref{gsub})$; {\textcolor{black}{spatio-temporal denoising/de-aliasing of the deformed scene}}; \\
       \hspace*{6em} $f_{n} \leftarrow$$(\ref{fsub})$; {CG reconstruction update with deformation correction};\\
       \hspace*{9em}  \textbf{if } ($\rm cost_{n}-\rm cost_{n-1}/\rm cost_{n}<10^{-3}$); cost as defined in $(\ref{maincost})$;   \\
        \hspace*{12em} break the inner loop;\\
       \hspace*{9em} \textbf{end}\\
       \hspace*{3em} \textbf{end}\\
       \hspace*{3em} \textbf{while} ($\|\theta_{n}-\theta_{n-1}\|_{2}^{2}/\|\theta_{n}\|_{2}^{2}>10^{-2}$); Check if the deformation parameter update has converged;\\
      \hspace*{6em} $\theta_{n}\leftarrow$$(\ref{thetasub})$; {Deformation estimation};\\
      \hspace*{3em} \textbf{end}\\
      \hspace*{3em} $\alpha = \alpha \times 3$; continuation over the deformation force strength parameter;\\
      \hspace*{3em} $\beta = \beta \times 10$; continuation over the penalty parameter;\\
         \textbf{     end}
      \begin{center}
\line(1,0){250}
\end{center}

\textcolor{black}{Note that the deformation compensated images ${\cal T}_{\theta} \cdot  f$ is the denoised version of $f$ only in the first iteration because the deformation estimate in the first iteration is zero. The deformation field $\theta_{1}$ estimated at the first iteration is used to derive the deformation compensated image in the second iteration; it is obtained as the denoised version of ${\cal T}_{\theta_{1}}\cdot f$. The iterations, along with the continuation of the $\beta$ parameters, will reduce the risk of the algorithm to converge to undesirable local minima. As the iteration number $n$ increases, we have ${\cal T}_{\theta_{n-1}}\cdot f_{n} \approx g_{n}$. }

\section{Experimental evaluation}
To validate the proposed DC-CS scheme, we perform retrospective resampling experiments based on ground truth data from (i) the Physiologically improved non-uniform cardiac torso (PINCAT) numerical phantom, \cite{sharif2007physiologically}, \cite{lingala2011accelerated}, (ii) an in vivo fully sampled myocardial perfusion MRI dataset with \textcolor{black}{breathing} motion. We compare the proposed DC-CS scheme with different choices of compactness priors ($\Phi$ in (\ref{maincost})) against: (a) CS schemes that use the same priors, and \textcolor{black}{(b) the k-t FOCUSS with ME/MC scheme}.  We  show example myocardial perfusion MRI reconstructions using  radial data from \textcolor{black}{two subjects} imaged during shallow breathing at stress.\textcolor{black}{ We finally demonstrate a scenario of accelerating an un-gated myocardial perfusion MRI dataset that contains both cardiac and respiratory dynamics in addition to the contrast dynamics.}
\subsection{Datasets}
\subsubsection{PINCAT data}
We numerically simulate the acquisition of myocardial perfusion MRI data from a single short axis slice of the PINCAT phantom \cite{sharif2007physiologically, lingala2011accelerated}. We set the phantom parameters to obtain realistic cardiac perfusion dynamics and contrast variations due to bolus passage, while accounting for respiration with variability in breathing motion. The contrast variations due to bolus passage are realistically modeled in regions of the right ventricle (RV), left ventricle (LV) and the left ventricle myocardium. A temporal resolution of one frame per heart-beat is assumed. The time series data consists of 35 time frames capturing the first pass passage of bolus through the different regions of the heart. The spatial matrix size is 64 x 64, which corresponds to a spatial resolution of 3 x 3 mm$^{2}$. For retrospective undersampling, we simulate the acquisitions assuming golden angle \textcolor{black}{pseudo} radial $k-t$ sampling pattern; (the angle between successive rays was 111.25 degrees). \textcolor{black}{The pseudo radial sampling involved gridding of the radially sampled data to the nearest point on a Cartesian grid.} Subsampling was performed by considering (30 to 8) rays/frame respectively.

\subsubsection{In vivo fully sampled ECG gated myocardial perfusion MRI data}
We consider a single slice from a fully sampled in vivo myocardial perfusion MRI scan. A healthy subject was scanned on a Siemens 3T Trio scanner at the University of Utah in accordance with the institutes review board. Data was acquired using a saturation recovery FLASH sequence (TR/TE=2.5/1ms, Saturation recovery time =100ms, 3slices). A Cartesian grid (phase encodes $\times$ frequency encodes: $90 \times 190$, temporal resolution: 1 beat, \textcolor{black}{spatial resolution: 2.5 mm $\times$ 2.5 mm $\times$ 8 mm}) and a Gadolinium bolus of 0.04mmol/kg was used under rest conditions. The data contains motion primarily due to breathing and inconsistent gating. We added additional integer shifts to amplify motion \textcolor{black}{(see fig \ref{cartcomparisononmccs})}. Similar to the numerical phantom, we used the golden angle \textcolor{black}{pseudo} radial sampling at different subsampling levels (30 to 12 rays/frame) for retrospective undersampling.

\subsubsection{Radial myocardial perfusion data sets}
Data was acquired using a perfusion radial FLASH saturation recovery sequence (TR/TE $\approx$ 2.6/1.2 ms, 3 slices per beat, \textcolor{black}{flip angle of 14 degrees, 2.3 $\times$ 2.3 $\times$ 8 mm pixel size, \textcolor{black}{FOV: 280 mm$^{2}$, bandwidth 1002 Hz/pixel })} on a Siemens 3T \textcolor{black}{Trio} scanner  \cite{dibella2011effect} using the Siemens cardiac coil array. 
 72 radial rays equally spaced over $\pi$ radians and with 256 samples per ray were acquired for a given time frame and a given slice. These rays were acquired in an interleaved manner in subsets of 6 rays each. The rays in successive frames were rotated by a uniform angle of $\pi/288$ radians, which correspond to a period of 4 across time.  We considered \textcolor{black}{two} stress data set\textcolor{black}{s} that were acquired on a free breathing normal subject, \textcolor{black}{and a patient with suspected ischemia} where 0.03 mmol/kg of Gd contrast agent was injected after 3 minutes of adenosine infusion. A PCA based coil compression strategy \cite{buehrer2007array} was used to compress the four coil data set to a single coil principal component data set. With this data, we performed single coil reconstruction comparisons using 24 rays that were chosen to \textcolor{black}{approximately follow the golden angle distribution.}
  
 \subsubsection{\textcolor{black}{In-vivo fully sampled ungated myocardial perfusion MRI data}}
\textcolor{black}{ We consider a single slice from a fully sampled un-gated  myocardial perfusion MRI scan. We considered running the same saturation recovery sequence (TR/TE=2.5/1ms, saturation recovery time = 100ms, 1 slice)   on a different subject but with no gating (phase encodes $\times$ frequency encodes: 108x190, temporal resolution: $\approx$ 4frames/beat,\textcolor{black}{spatial resolution:  2.5 mm $\times$ 2.5 mm $\times$ 8 mm}). The subject breathed heavily during this scan. The data set was subsampled using 30 rays /frame using the pseudo-golden angle radial sampling patterns.  Note that the data contains pseudo cardiac motion on the top of breathing motion. Note from the x-t profile in Fig. \ref{ungatedcartcomparisononmccs} (fifth column) and its temporal Fourier transform (sixth column) that the energy is distributed at almost all frequencies, mainly due to the modulation of pseudo-periodic cardiac motion by respiratory motion; this limits the utility of CS schemes with x-f sparsity. It is also challenging for the registration algorithm to register frames in different cardiac phases to obtain smooth x-t profiles. Specifically, the smoothness constraints on the motion field have to considerably relaxed, which will result in a less stable algorithm. We capitalize on the ability of the DC-CS scheme to use Fourier sparsity penalty since pseudo-periodic cardiac motion will still be sparse in the Fourier domain, while keeping the smoothness constraints on the deformations to be the same as in experiments for free breathing gated data. This will correct for bulk respiratory motion, but have limited ability in correcting for cardiac motion. }


\begin{figure*}

\centering
{\includegraphics[width=1\textwidth]{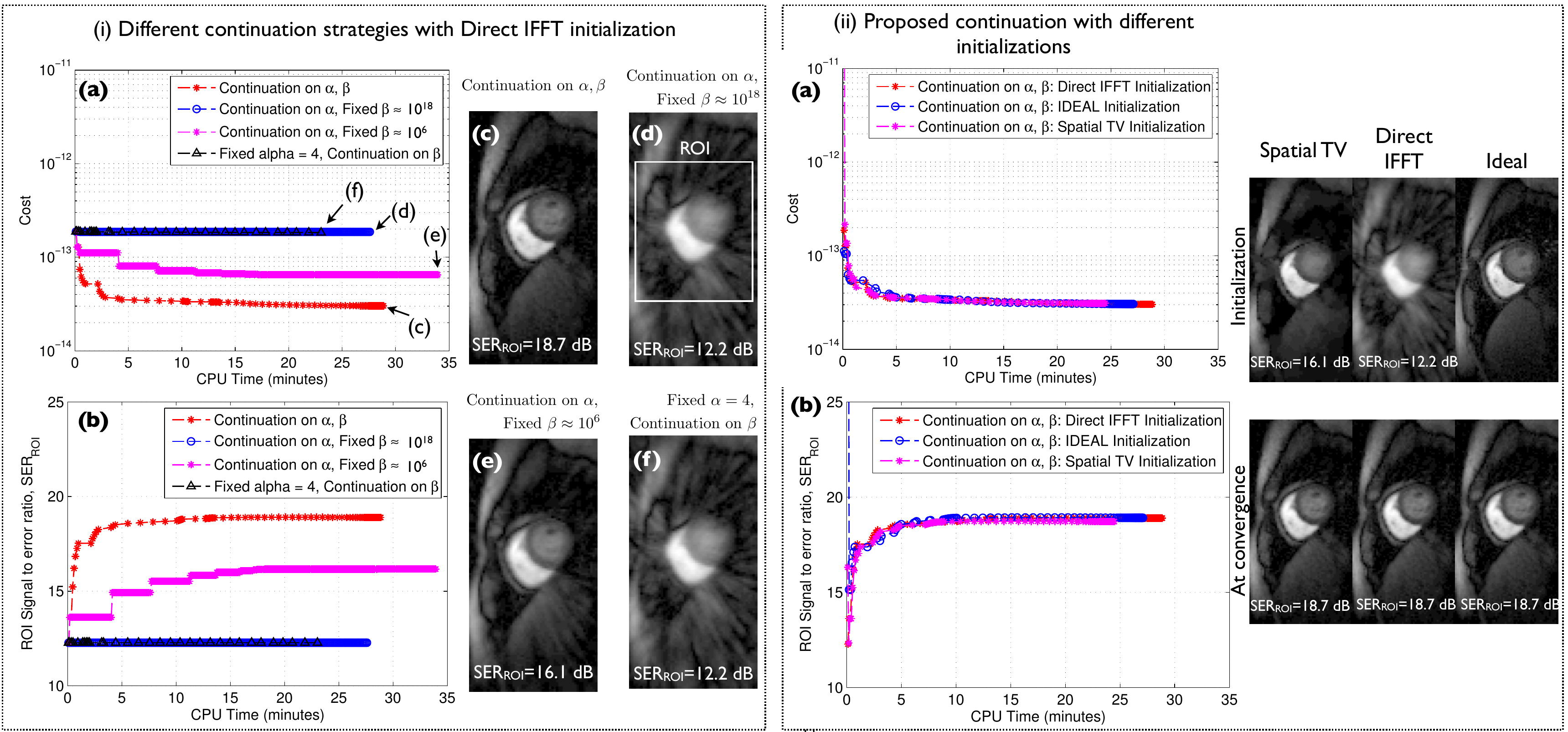}}
 \caption{ 
 Role of continuation: In (i), we study the evolution of the cost in $(\ref{maincost})$ and the $\rm SER_{\rm ROI}$ as a function of the CPU run time using different continuation strategies with the zero filled direct IFFT reconstruction as the initialization. Here, we compare the proposed continuation strategy of updating $\alpha,\beta$ against other strategies of not performing continuation on either one of the two parameters $\alpha, \beta$. It can be seen that the scenarios without continuation depict poor convergence and result in undesirable solutions (see i.(d,e,f)). In contrast, it can be seen from i.(a,b,c) that the continuation over both $\alpha, \beta$ depicts improved convergence and obtains a solution in i.(c) that is free of artifacts and blur. Further in (ii), we show the cost in $(\ref{maincost})$ and $\rm SER_{\rm ROI}$ versus CPU run time with the proposed continuation scheme for different initializations. It can be seen that the algorithm converged to approximately the same solution irrespective of the choice of the initialization; the robustness to the initialization is attributed to the continuation rules which ensures a gradual update of the complexity of the problem.}
\label{initguessplot}
\end{figure*}

 \begin{figure}
\centering
{\includegraphics[width=0.481\textwidth]{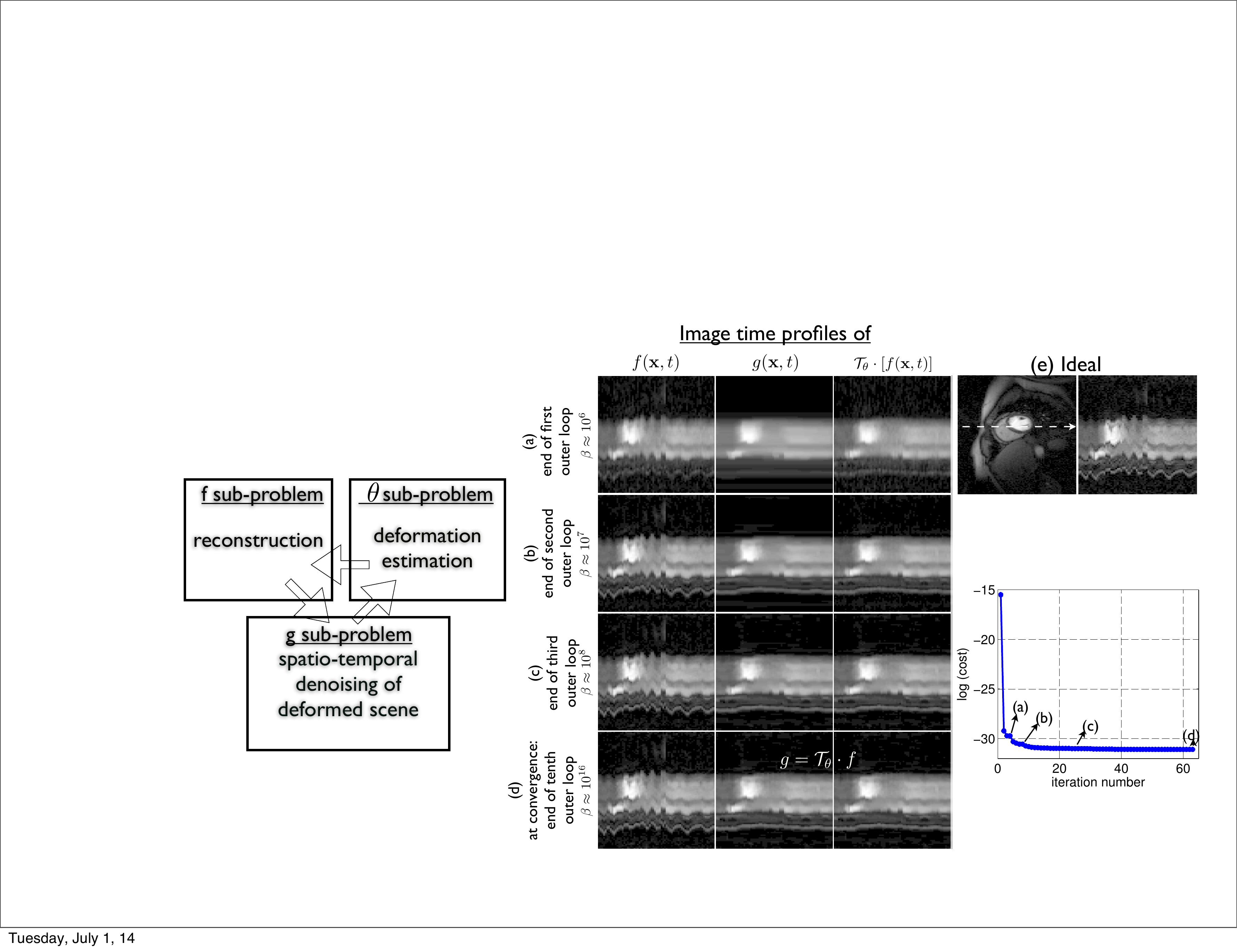} }

\caption{\textcolor{black}{Evolution of solutions in different sub-problems: In this schematic, we show the time profiles of $f,g,{\cal T}_{\theta}\cdot f$ as they evolve during the iterations. From the first row, it can be seen that in the initial iterations, the value of $\beta$ is small that resulted in the  smooth reference dataset in $g$. This dataset is image registered frame by frame to the reconstruction $f$ to obtain the motion estimates $\theta$. Note that $g$ and ${\cal T}_{\theta}\cdot f)$ are not too similar in the first row. Specifically, since $\beta$ is small, the constraint ${\cal T}_{\theta}\cdot f=g$ is not satisfied. As the algorithm converges, we increment $\beta$ in a continuation manner, and hence the constraint in eq. (\ref{const}) $(g={\cal T}_{\theta}\cdot f)$ is satisfied at convergence.  Also, note that during each iteration the reference scene is obtained as the denoised version of the motion corrected dataset ${\cal T}_{\theta}\cdot f. $}  }
\label{continuationimages}
\end{figure}

\subsection{Metrics used for quantitative comparison}
In the retrospective undersampling experiments, we evaluate the performance of different methods in terms of the signal to error ratio (SER) metric and the high frequency signal to error  (HFSER) metric. These metrics are evaluated within the field of view that contained the regions of the heart. This was motivated by recent findings in \cite{bilen2010compressed}, and by our own experience in determining a quantitative metric that best describes the accuracy in reproducing the perfusion dynamics in different regions of the heart, and the visual quality in terms of preserving crispiness of borders of heart, and minimizing visual artifacts due to reconstructions. The metrics are defined below where $\rm ROI$ denotes a square field of view containing regions of the heart (see figures (\ref{pincatcomparisonmccs}), (\ref{cartcomparisononmccs}) for the $\rm ROIs$ used in this paper; \textcolor{black}{these were heuristically drawn on the fully sampled dataset such that the heart always lie within the ROI across all the frames.}):

\begin{figure*}
 \centering
{\includegraphics[width=0.8\textwidth]{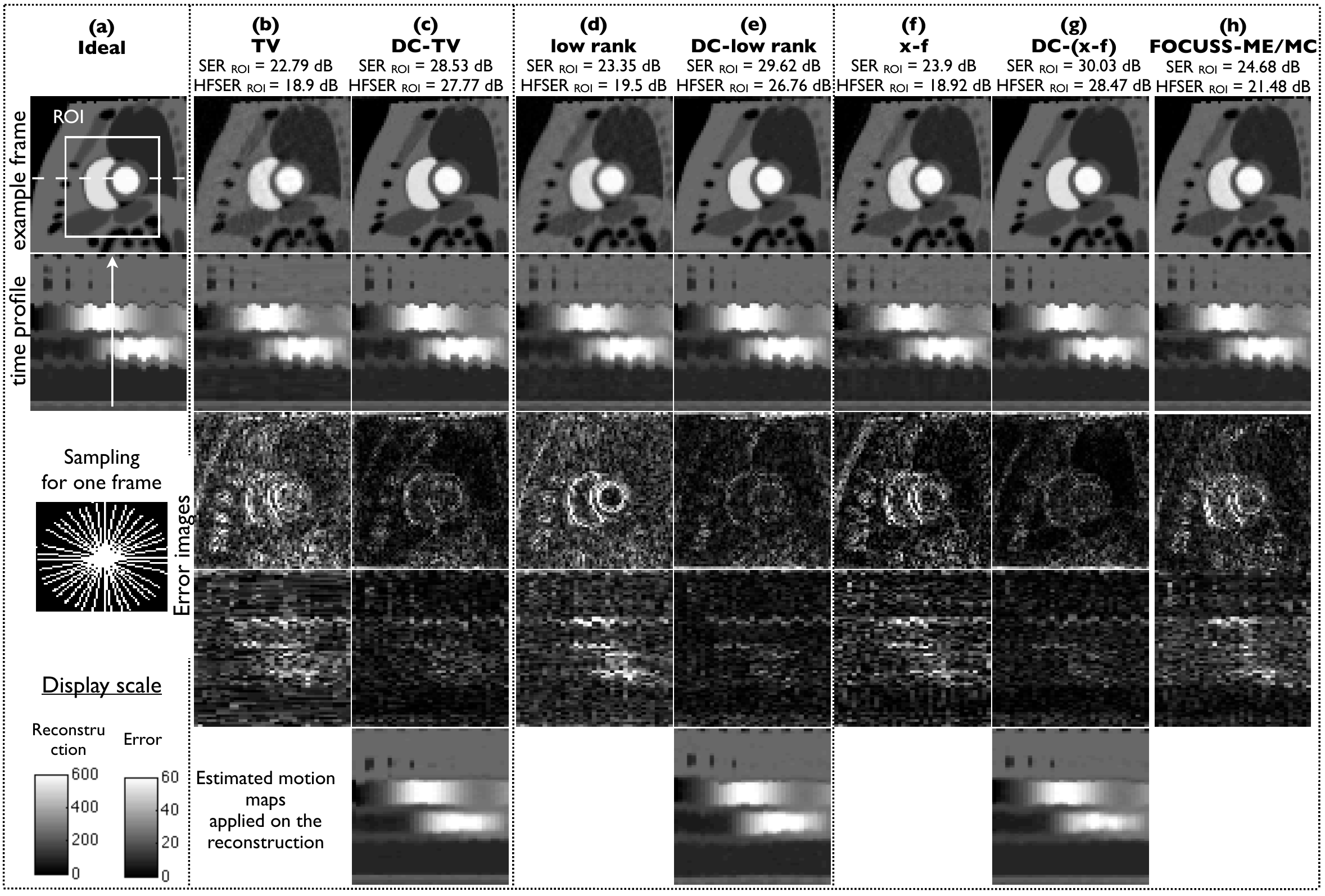}} 
\caption{Qualitative comparison on the PINCAT phantom data using 20 rays/frame: We show for each of the scheme\textcolor{black}{s} a spatial frame, its image time profile, and the corresponding \textcolor{black}{difference} images and \textcolor{black}{difference} time profiles. The \textcolor{black}{difference} images are scaled up by a factor of 10 for better visualization. The time profile in the second row corresponds to the profile along the dotted line in the first image frame of the top row. The location of the image frame is depicted by the arrow in the time profile of (a).  \textcolor{black}{As depicted from these figures,  DC-CS  provides superior reconstructions compared to CS with regards to reduced motion artifacts}. The motion compensated time profiles shown in the bottom row depicts that the proposed algorithm was capable to estimate and correct most of the inter-frame motion. }
\label{pincatcomparisonmccs}
\end{figure*}
\textcolor{black}{Given a reconstructed data set $\gamma_{\rm recon}(\mathbf x,t)$, and a ground truth dataset $\gamma_{\rm ideal}(\mathbf x,t)$, the SER and HFSER are defined as:} 
\begin{equation}\label{SER}
{\rm SER}_{\rm ROI} = -10\log_{10}\frac{1}{N}\sum_{i=1}^{N}{\frac{\| (\rm  \gamma_{recon,i}) -(\rm  \gamma_{ideal,i})\|_{F}^{2}}{\| (\rm  \gamma_{ideal,i})\|_{F}^{2}}};
\end{equation}
where $N$ is the number of time frames. The {\rm HFSER} metric which gives a measure of image sharpness is given by \cite{ravishankar2011mr, lingala2013blind}: 
 \begin{equation}
{\rm HFSER}_{\rm ROI} = -10\log_{10}\frac{1}{N}\sum_{i=1}^{N}{\frac{\| \rm LoG( \gamma_{recon,i}) - LoG( \gamma_{ideal,i})\|_{F}^{2}}{\| \rm LoG( \gamma_{ideal,i})\|_{F}^{2}}};
\end{equation}   
where $\rm LoG$ is  a Laplacian of Gaussian filter that capture edges. We use the filter specified by a kernel size of $15 \times 15$ pixels, with a standard deviation of $1.5$ pixels.

\subsection{Implementation}
All the methods were run on a linux machine with an \textcolor{black}{Intel Xeon CPU processor (3.6 GHz, 8 cores) and a 31.4 GB of RAM}. Both DC-CS and CS reconstructions were optimized by tuning the corresponding regularization parameter $\lambda$ in $(\ref{maincost})$ that gave the maximal $\rm SER_{\rm ROI}$, specified by $(\ref{SER})$. The CS reconstructions were implemented by considering ${\cal T}={\cal I}$ in $(\ref{maincost})$. We used the initial values of $\alpha=4$ and used the same continuation rule updates for $\beta, \alpha$ for all the three compact priors in DC-CS. The value of the standard deviation $\sigma$ of the Gaussian regularizer in the demons algorithm was fixed to a standard deviation of 10 pixels for all the compact priors. \textcolor{black}{It is to be noted that a larger choice of the Gaussian kernel will restrict the amount of deformation that could be corrected, on the other hand a smaller size of the kernel can get stuck in an undesirable local minima. We chose the kernel size ($\sigma=6$) heuristically such that the bulk breathing motion is corrected.}
\textcolor{black}{The optimal tuned value of $\lambda$ for the DC-CS schemes were found to be lower than that of the CS schemes for all the datasets. For different CS priors, the value of $\lambda$ was slightly higher with the nuclear norm prior than that of the sparsity priors. For the DC-CS scheme, the tuned $\lambda$ did not vary much across the different priors.  }
For the radial data, where we did not have fully sampled truth images,  we chose the regularization parameter for all the methods based on the $L$-curve heuristic;  the choice of the regularization parameter from this heuristic also empirically matched the reconstructions that depict the best image quality. 




\begin{figure}
\centering
\subfigure {\includegraphics[width=0.241\textwidth]{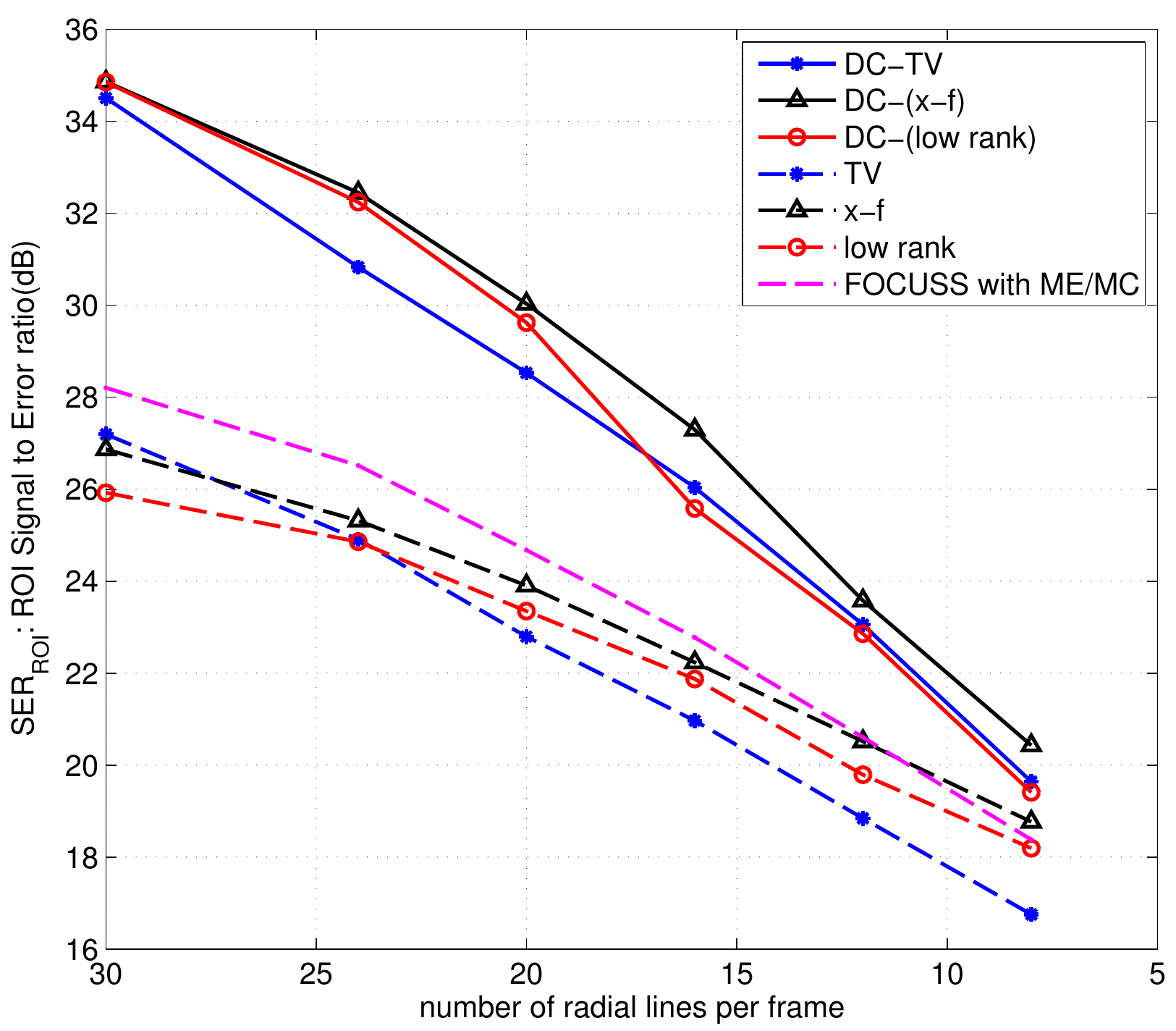}} 
\subfigure {\includegraphics[width=0.241\textwidth]{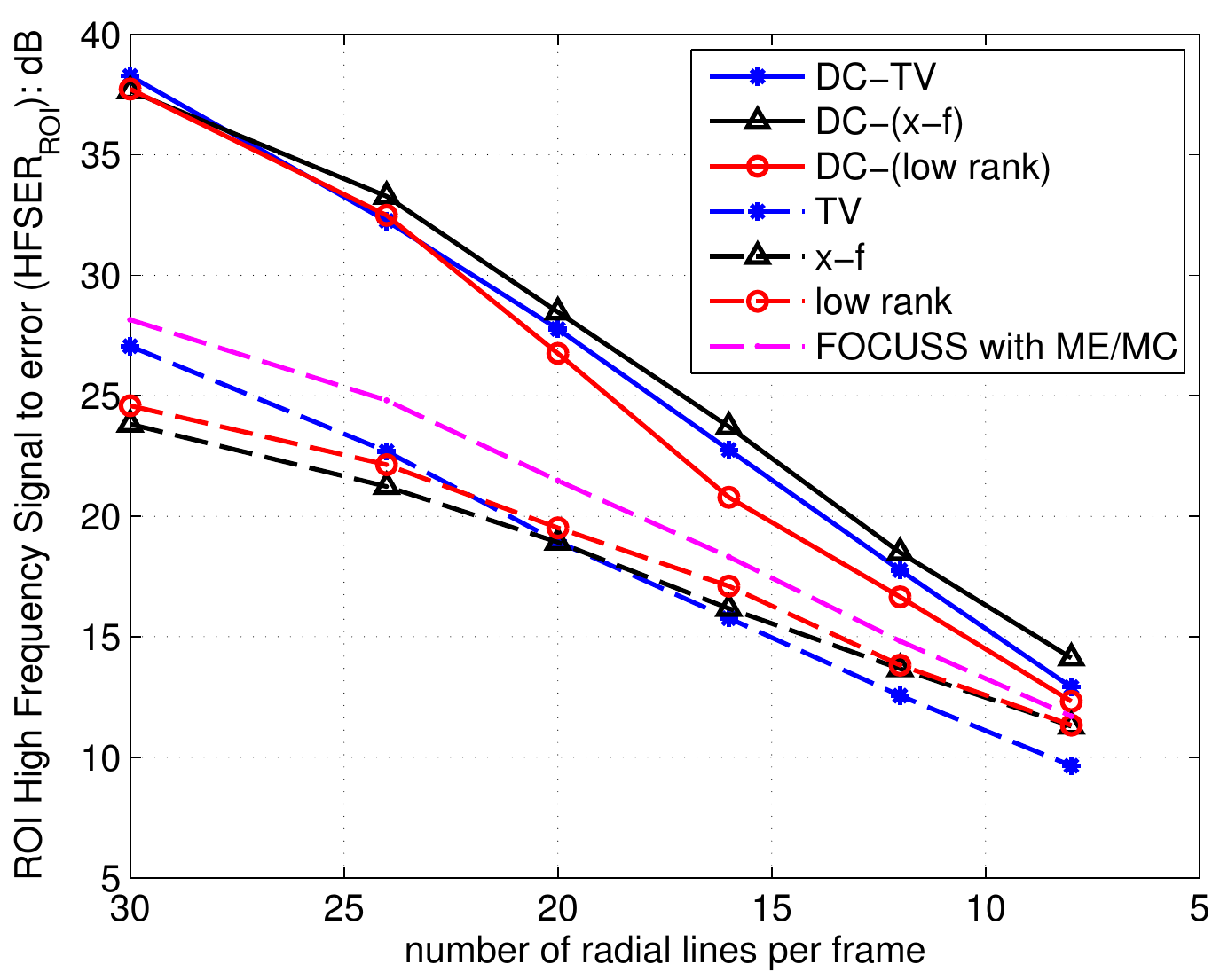}} 
\caption{Quantitative comparisons of different schemes using  (a) the signal to error ratio, and the (b) high frequency signal to error metrics on PINCAT phantom data. These metrics are evaluated in a square field of view that contains regions of heart as depicted in Fig. \ref{pincatcomparisonmccs}. These plots demonstrate that the DC-CS schemes  outperform the CS schemes at all sub-sampling factors.}
\label{SERandHFENpincatmccs}
\end{figure}



\subsection{Convergence analysis}

In this section, we study the convergence behavior of the proposed DC-CS algorithm. We demonstrate the convergence properties by considering the in vivo myocardial perfusion MRI dataset. We consider the recovery of this data from Fourier golden angle undersampled radial data using 16 rays/frame; which approximately correspond to an acceleration factor of 5 . We employ $\Phi$ to be the temporal finite difference (temporal TV) operator. We now demonstrate the role of the continuation, and discuss the algorithm's dependence on the initial guess of the reconstruction. 

\subsubsection{Role of continuation}
 As detailed in the pseudo code above, we define continuation over the deformation force strength parameter ($\alpha$) and the penalty parameter ($\beta$) as a strategy of  starting with small values of $\alpha, \beta$ and incrementing them by small factors in an outer loop. In order to evaluate the role of continuation, we study the below scenarios in Fig. \ref{initguessplot}(i) with an initial guess determined by the zero filled  direct Inverse Fourier Transform (DIFFT) reconstruction; here, we  study the decrement of the cost function in $(\ref{maincost})$ and the  evolution of the $\rm SER_{\rm ROI}$ (\ref{SER}):
\begin{itemize}
\item Proposed strategy of continuation over both $\alpha, \beta$ starting with  $\alpha=4, \beta \approx 10^{6}$;
\item Continuation over $\alpha$ starting with $\alpha=4$ with a fixed low value of $\beta \approx 10^{6}$;
\item Continuation over $\alpha$ starting with $\alpha=4$ with a fixed high value of $\beta \approx 10^{18}$;
\item Fixed low value of $\alpha=4$ with a continuation over $\beta$;
\end{itemize}
\textcolor{black}{A fixed high value of $\alpha$ was not considered, as this means that the strength of deformation forces are almost close to zero, resulting in  minimal to no motion correction.}
\textcolor{black}{Note the initial values of $\beta$ above are derived as $\beta_0=\frac{1}{max(\Phi({\cal T}_{\theta_{init}}\cdot f_{init}))}$, and an empirical choice of $\alpha_{0}=4$. }
From Fig. \ref{initguessplot} (i), we observe that the scenarios without the use of continuation had poor convergence properties. Specifically, the scenario of using a high value of $\beta$ showed slow convergence \textcolor{black}{irrespective of the continuation over $\alpha$ (due to many CG steps while solving the f sub-problem)}, while a low value of $\beta$ showed fast convergence but resulted in an in-accurate solution. We also observed slow convergence without continuation over $\alpha$. In contrast,  we observe that the proposed strategy of continuation over $\beta, \alpha$ to show both improved convergence speed and achieve the desired solution. \textcolor{black}{We refer the reader to Fig. \ref{continuationimages} to get a sense of the outputs of $f,g,{\cal T}_{\theta}\cdot f$ during the iterations of the DC-CS scheme with continuation.}


\subsubsection{Choice of initial guess}: In Fig. \ref{initguessplot} (ii), we study the  behavior of the algorithm with the proposed continuation scheme to different initial guesses of $f$. We consider different initial guesses obtained from the fully-sampled ground truth data, zero filled  direct IFFT reconstructed data, and a spatially regularized total variation (TV) reconstructed data. From  Fig. \ref{initguessplot} (ii), we observed that that the algorithm was robust to the choice of the initialization. The continuation ensures a gradual progression in the complexity of the problem, thereby avoiding the chance of getting stuck in undesirable local minima. Based on these observations, we utilized the choice of the spatial total variation (TV) reconstruction as the initial guess to the proposed DC-CS scheme for the rest of the experiments in the paper.

\begin{figure*}
\centering
{\includegraphics[width=0.75\textwidth]{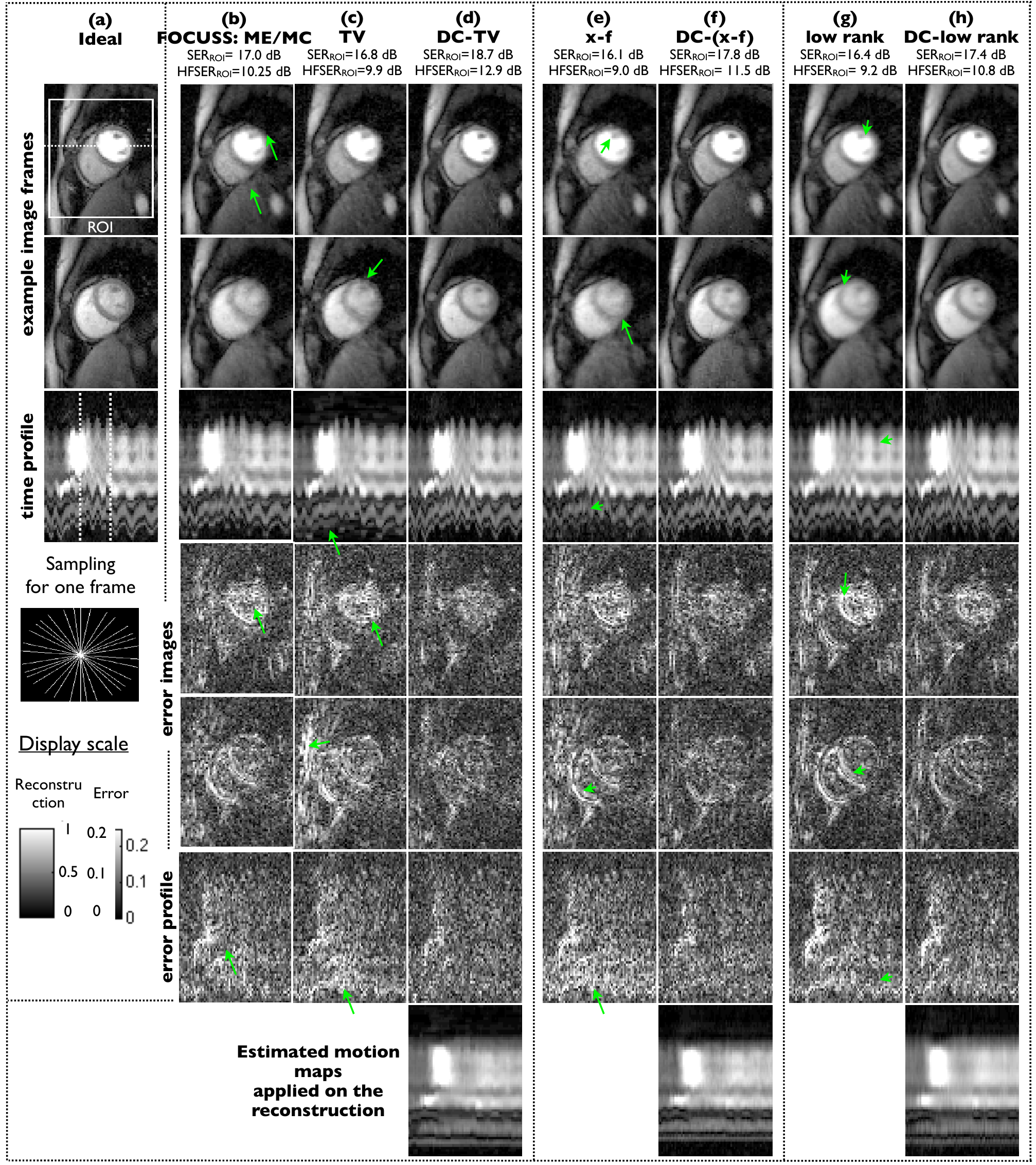}} 
\caption{Performance evaluation using retrospective sampling on Cartesian data using 16 radial rays/frame: We show two example frames,  the image time profile along the dotted line in (a), and the corresponding \textcolor{black}{difference} images for all the methods. The \textcolor{black}{difference} images are scaled up by a factor of five for better visualization. The image time frame locations in the first two rows are depicted by the dotted lines in the time profile of (a). The compressed sensing (CS) reconstructions exhibit considerable motion artifacts and temporal blurring (see arrows), while the proposed deformation corrected CS images (DC-CS) are robust to these compromises. The last row depicts that the proposed algorithm was capable of estimating and correcting most of the deformation due to inter frame motion.  }
\label{cartcomparisononmccs}
\end{figure*}

\subsection{Comparisons on the PINCAT numerical phantom}

In Fig. $\ref{pincatcomparisonmccs}$, we show qualitative comparisons of the proposed DC-CS scheme with different choices of {potential functions/compactness priors} against CS schemes that use the same priors. The qualitative comparisons are shown by considering undersampling using 20 rays/frame. We observe that all the three priors benefit from deformation/motion correction. Specifically, classical CS methods result in temporal stair casing (with temporal TV), motion blurring (with temporal Fourier), loss in spatio-temporal fidelity and blurring of myocardial borders (with the nuclear norm based low rank prior). In contrast, the proposed DC-CS methods are found to be more robust to these artifacts. We observe similar trends over a range of subsampling factors as depicted in the $\rm SER_{\rm ROI}$ and ${\rm HFSER_{\rm ROI}}$ plots in Fig. $\ref{SERandHFENpincatmccs}$.

\subsection{Comparisons on the fully sampled  invivo myocardial perfusion MRI dataset}

\begin{figure}
\centering
\subfigure {\includegraphics[width=0.241\textwidth]{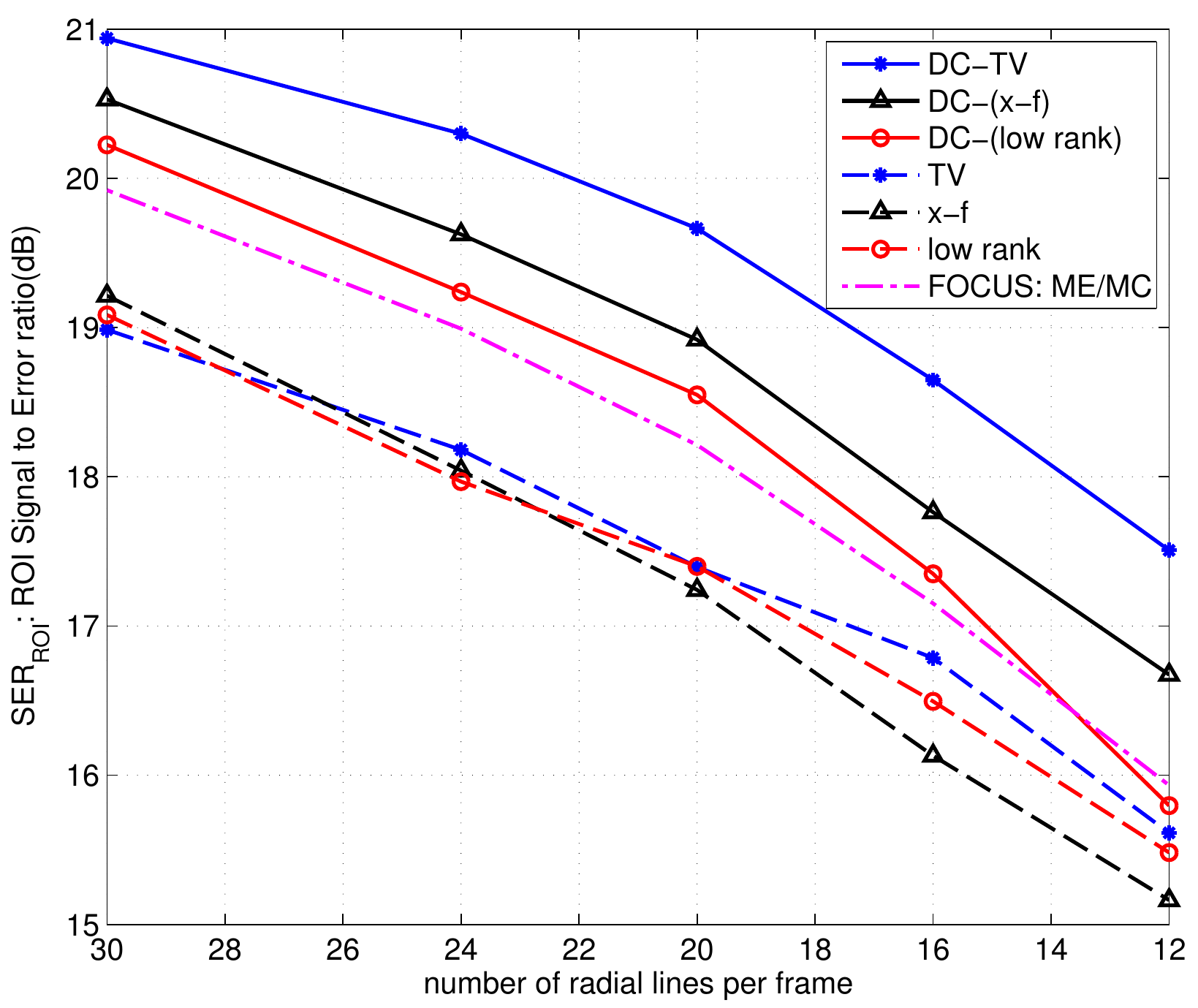}} 
\subfigure {\includegraphics[width=0.241\textwidth]{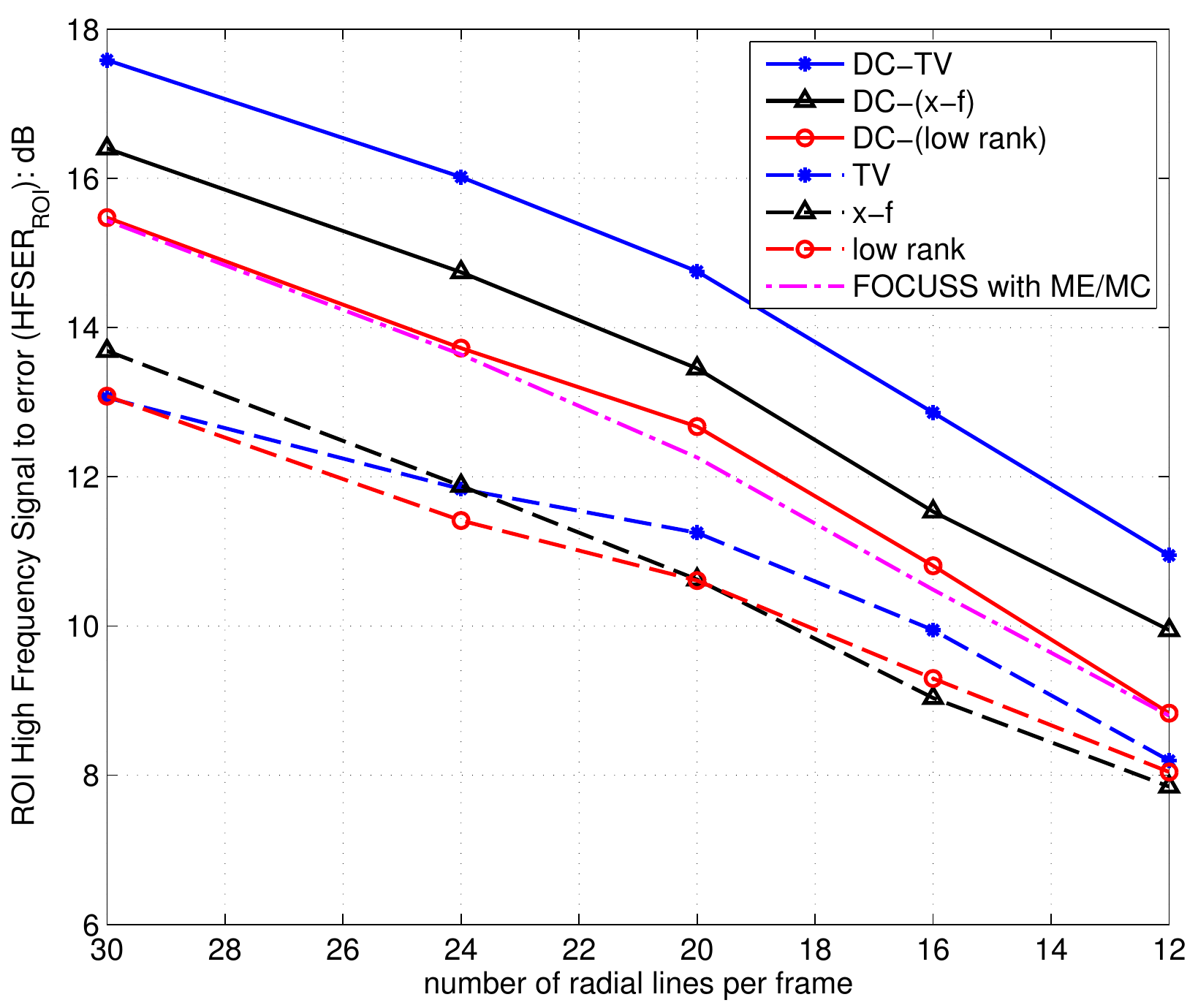}} 
\caption{Quantitative comparisons of different schemes using the signal to error ratio (a), and high frequency Signal to error  metrics (b) on the in vivo myocardial perfusion data. The metrics are evaluated in the regions of interest containing the heart as depicted in Fig. \ref{cartcomparisononmccs}. These plots demonstrate that the DC-CS schemes  outperform the CS schemes at all sub-sampling factors. }
\label{SERandHFENrealmccs}
\end{figure}

In Fig. $\ref{cartcomparisononmccs}$, we show the comparisons involving retrospective sampling on the fully sampled in vivo Cartesian data. The comparisons are shown using 16 rays per frame. Similar to the PINCAT phantom observations, we notice superior spatio-temporal fidelity and less motion artifacts with the proposed DC-CS scheme compared to its CS variants. The $\rm SER_{\rm ROI}$ and ${\rm HFEN_{\rm ROI}}$ plots in Fig. $\ref{SERandHFENrealmccs}$ also depict the same trend over a range of subsampling factors.

\subsection{Comparisons on radial shallow breathing stress data}

\begin{figure*}
{\includegraphics[width=1\textwidth]{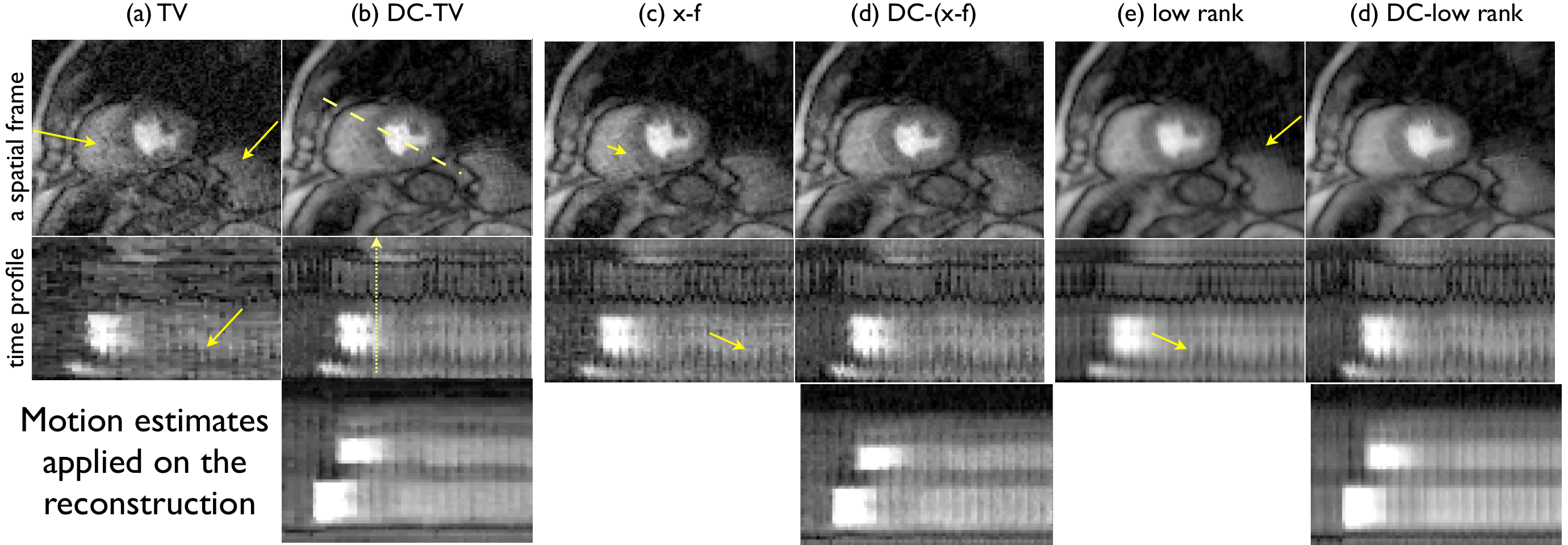}} 
\caption{Comparison on undersampled shallow breathing radial data using 24 rays/frame. Here single coil reconstructions are shown. For all the methods, an example frame is shown in the top row while the time profiles along the dotted line in (b) is shown in the bottom row. It can be seen that DC-CS had fewer motion artifacts than CS schemes (see the arrows along the myocardial borders that depict motion blur and temporal blurring in CS). }
\label{prospectivecomparisonsmccs}
\end{figure*}

\begin{figure*}
\centering
{\includegraphics[width=0.81\textwidth]{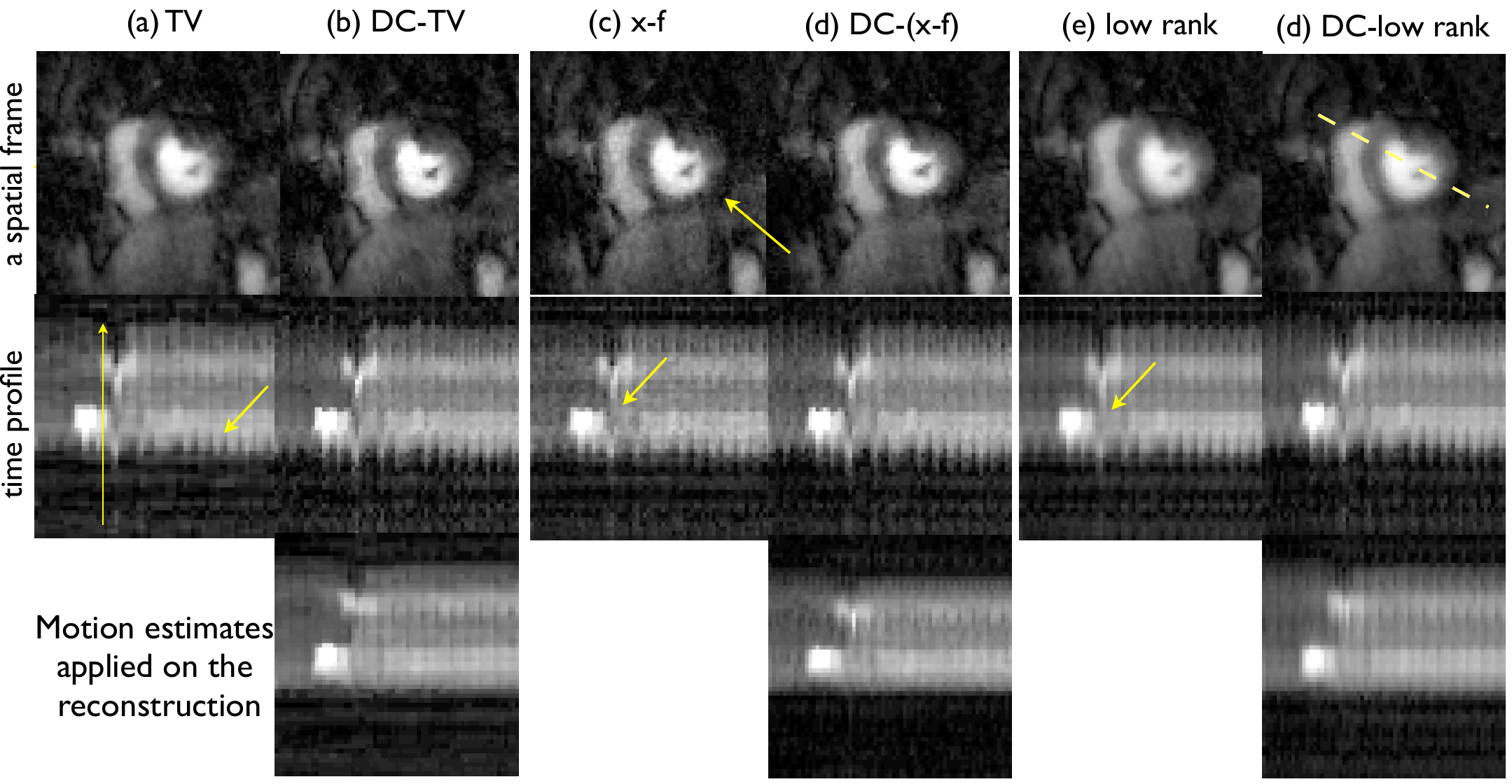}} 
\caption{\textcolor{black}{Comparison of under sampled shallow breathing radial data using 24 rays/frame on a patient with suspected ischemia: Similar to Fig.8, single coil reconstructions are shown. It can be seen that the DC-CS schemes show better fidelity in terms of reduced motion artifacts over CS schemes (see arrows that depict motion blurring in the CS scheme).}}
\label{patient1stress}
\end{figure*}

\begin{figure*}
\centering
{\includegraphics[width=0.81\textwidth]{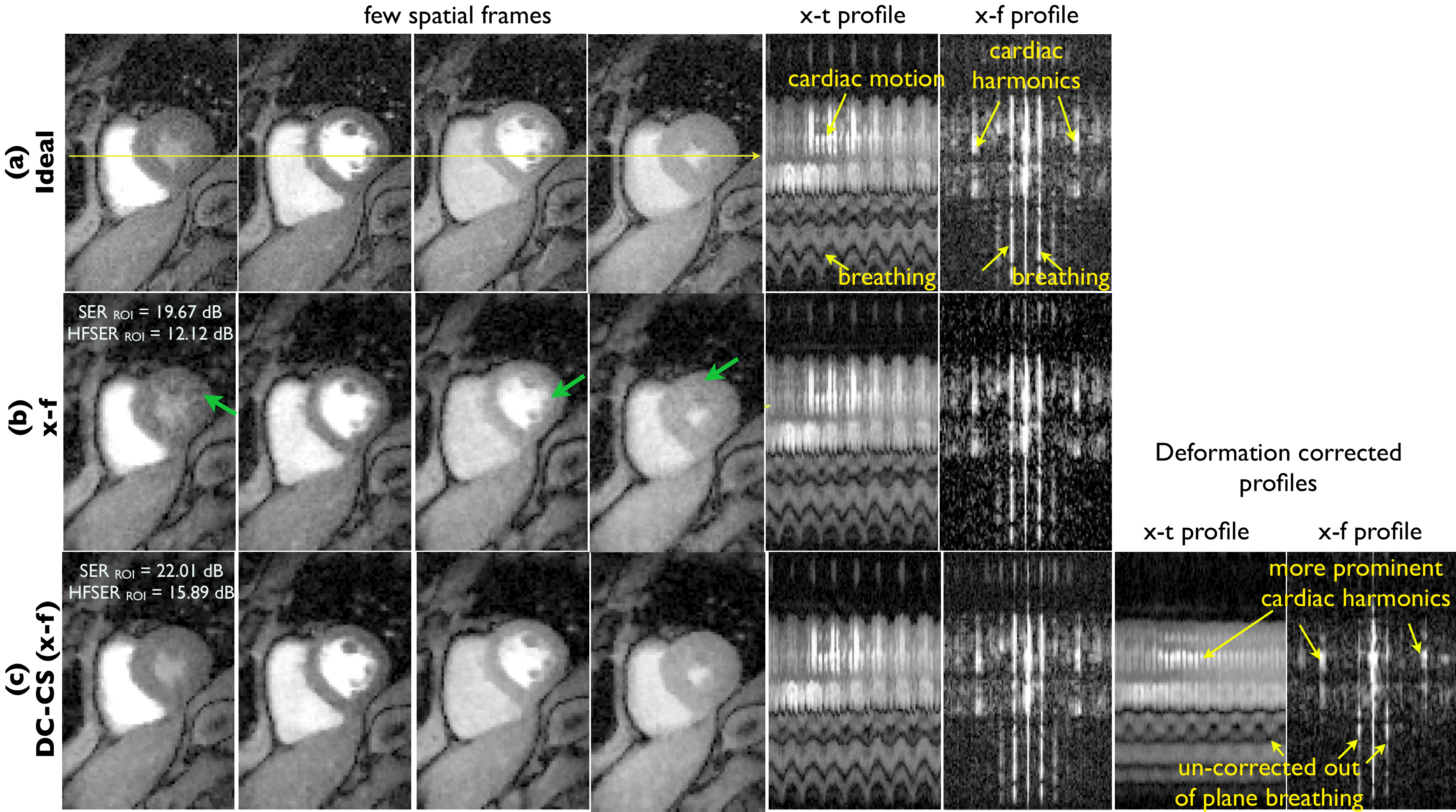}}
\caption{\textcolor{black}{Performance evaluation using retrospective undersampling of ungated perfusion MRI dataset: The ungated dataset was acquired without ECG gating and breath-holding as described in Section IV.A;  A few frames of the original data, data recovered using CS with x-f sparsity penalty, and the DC-CS scheme are shown in the first four columns. The ungated acquisition enables us to acquire diastolic and systolic frames. However, the acceleration of this dataset is challenging due to the rapid cardiac motion (see high frequency ripples in x-t profile) and respiratory motion (see low frequency oscillations in x-t profile). The x-f space representation of the ground truth data depicts the modulation of the cardiac and contrast dynamics by the breathing motion at almost all the frequencies. The proposed DC-CS scheme corrects for the breathing motion, and exploits the sparsity of quasi-periodic dataset with cardiac and contrast dynamics (see the x-f profile of the motion corrected dataset in the last row). The residual ripples in the x-t profile of the deformation corrected dataset correspond to out of plane breathing motion. From the x-f, and x-t plots, it can be seen that DC-CS  provides superior reconstructions over CS reconstructions (also see arrows in (b)). 
}}
\label{ungatedcartcomparisononmccs}
\end{figure*}

In Figs. $\ref{prospectivecomparisonsmccs}$, and $\ref{patient1stress}$, we show the comparisons using undersampled radial stress shallow breathing data from \textcolor{black}{two subjects}. We observe DC-CS to give better reconstructions in terms of minimizing motion blur and artifacts compared to CS. These preliminary results are demonstrated using a single coil with 24 rays, however the performance could be  improved by extending to multiple coils and including spatial priors.  Without the motion compensated, the low rank prior was generally robust to motion artifacts but sensitive to temporal blurring during the peak contrast frames while the CS priors based on temporal TV and temporal FFT were sensitive to motion artifacts. With the DC-CS scheme, we observe the reconstructions with temporal TV and temporal Fourier to be slightly superior to that of the low rank prior in terms of image sharpness and reduced blurring. We however realize that the performance of the low rank priors can be improved by considering non-convex Schatten p-norm priors and/or patch based low rank priors, as shown in \cite{lingala2011accelerated,trzasko2011local}.

\textcolor{black}{\subsection{Comparisons on the ungated free breathing myocardial perfusion dataset}
We observe that the proposed algorithm provides less spatial and temporal blurring than the CS with x-f scheme without motion compensation, which can also be appreciated from the x-t profiles and x-f profiles. Note from the x-f profile of the deformation corrected dataset that the breathing motion corrected dataset is more sparse in the temporal Fourier domain. Specifically, the energy at the cardiac harmonics are increased, while the energy at the other frequencies are decreased. We also observe that the motion compensation attenuates the harmonics at the respiratory motion frequency, except for the out of plane motion that our motion compensation algorithm cannot compensate. The superiority of image quality in DC-CS over CS can also be seen in the $\rm SER_{\rm ROI}$ and $\rm HFSER_{\rm ROI}$ numbers. 
}







\section{Discussion}
In this paper, we proposed a generalized deformation corrected compressed sensing framework for dynamic MRI. The proposed formulation is general enough to handle a wide class of compactness priors. Example priors based on sparsity in the temporal Fourier transform, sparsity in the temporal gradient, low rank priors were considered in this paper. We developed a variable splitting based optimization algorithm to decouple the problem to multiple well understood subproblems.  We observed that all the priors benefited from the proposed DC-CS scheme when compared to the classical CS schemes that utilized the same priors. \textcolor{black}{In this work, we have observed that the global low rank prior in the form of nuclear norm provided inferior reconstructions in comparison to the temporal finite difference, and temporal Fourier transforms. The performance of the low rank prior can be improved by  using non-convex relaxations of the rank, such as the Schatten p-norm $(p<1)$\cite{goudaccelerated}, or by directly solving the rank minimization problem using greedy approaches \cite{haldar2009rank}. Furthermore, the performance of the low rank prior has shown to be considerably improved when combined with sparsity priors such as finite difference, and temporal Fourier Transform priors\cite{lingala2013accelerating,zhao2012image}. }

Since the proposed framework decouples the denoising/de-aliasing as a stand-alone problem, it is straightforward to consider \textcolor{black}{spatial regularizers such as wavelet, spatial total variation, or combine the benefits of different regularizers such as patch based low rank priors, or combining the low rank priors and the temporal Fourier or the the temporal TV sparse priors.} Furthermore, extensions to include $\ell_{p}; p<1$ norms is also possible by utilizing efficient shrinkage rules during $\ell_{p}; p<1$ minimization \cite{hu2012fast}. \textcolor{black}{The reconstruction problem can also be readily adapted to include information from multiple coils.}

\textcolor{black}{During the deformation estimation step (\ref{thetasub}), we utilized a spatial smoothness constraint on the deformation field to ensure well-posedness. In this work, since we considered regular free breathing datasets, we used the same spatial smoothness parameter of $\sigma=10$ while registering all the frames. The algorithm in theory can capture sudden jumps in motion such as patient gasps by controlling the spatial smoothness parameter at the corresponding frames. Furthermore, in regular breathing patterns, the framework can be improved by constraining the deformation field to be spatially, and temporally smooth. }

The algorithm was observed to be robust to the choice of initial guess. This is attributed to the continuation strategies we adapt to update our reconstructions and deformations, and the undersampling behavior of the radial pattern. \textcolor{black} {We observed that the spatial TV initialization provided a slight advantage in terms of convergence speed over direct IFFT and ground truth initializations, which motivated our choice of using spatial TV as an initialization in all our experiments }. In this work, we did not evaluate the proposed method with different choices of sampling patterns. For example, a 1D Cartesian pattern could result in  overlap artifacts, which may not serve as a good initialization. A natural way to get the algorithm working with such sampling pattern is to initialize it with a basic regularized reconstruction that is free of artifacts such as the spatial TV reconstruction used in this work.

\textcolor{black}{In this work, we do not consider to account for out of plane motion due to simplicity. However, during free breathing, out of plane motion is inevitable. Most of the current acquisition schemes in myocardial perfusion MRI are 2-D in nature; it is difficult to compensate for out of plane motion in these cases. Conceptually, the proposed scheme can be readily extended to 3-D, which when used with 3-D acquisition schemes can provide improved reconstructions. In addition, we also  observe that the DC-CS algorithm does not require perfect compensation of motion. Any residual motion will only result in a more complex image, which will be captured by the appropriate sparse/low-rank prior.}

\textcolor{black}{The run time of the entire algorithm was about 25 mins for datasets of size $190\times90\times70$. However, the current implementation was not optimized for speed. The main bottle neck of the reconstruction time was the deformation algorithm, which was implemented in MATLAB. For instance, the  computational time spent for registering a single frame is of the order of 5-6 seconds for a 190x90 matrix. Perfusion datasets typically contain about 50-60 time frames. The image registration algorithm was called for  three times during the iterations. Overall, the computational time for the registration algorithm was about 16 minutes for the 190x90x70 matrix depicted in figure 2. This clearly formed the crux of the algorithm. The algorithm could be further optimized by other state of the art optimized for speed deformable registration algorithms, and also by the usage of graphical processing units (GPUs).} \textcolor{black}{In this work, we employed the demons algorithm to approximately solve (\ref{thetasub}). This approximation is a slight limitation of our current DC-CS implementation. We will investigate the use of registration algorithms that directly minimize (\ref{thetasub}) and study the impact of the assumptions of the registration algorithm on the DC-CS framework in our future work. } \textcolor{black}{In our work, we did not constrain the deformation to be invertible. Future work also include posing invertibility constraints on the deformation; this can benefit the stability of the registration process as shown by \textcolor{black}{\cite{chun2010sufficient}.}  }

The current  radial invivo results reported in this work were evaluated only from \textcolor{black}{two} subjects. Future work of considering datasets from multiple patients is required to fully evaluate the clinical utility of the proposed method. 

\section{Conclusion}
We introduced a novel  deformation corrected compressed sensing algorithm for accelerated dynamic MRI. The proposed framework has a generalized formulation capable of handling a wide class of compactness/sparsity priors on the deformation corrected dynamic signal.  We developed an efficient variable splitting based optimization framework to decouple the complex joint reconstruction and deformation estimation problem to simpler problems of shrinkage based denoising, deformable registration, and quadratic optimization. The efficient decoupling of the subproblems makes the proposed scheme applicable to a wide range of dynamic MRI applications including dynamic contrast enhanced MRI applications. Efficient continuation strategies were devised to cycle between the subproblems. We demonstrated that the proposed algorithm with continuation was robust to choice of initialization.  Our experiments on \textcolor{black}{a numerical phantom and, in vivo myocardial perfusion MRI datasets} demonstrated that the proposed scheme was able to  reduce motion artifacts and temporal blurring that were \textcolor{black}{present} in compressed sensing reconstructions. 

\bibliographystyle{IEEEtran} 
\bibliography{refs_modified}

\end{document}